\documentclass[10pt,journal,compsoc]{IEEEtran}
\ifCLASSOPTIONcompsoc
  \usepackage[nocompress]{cite}
\else
  \usepackage{cite}
\fi
\ifCLASSINFOpdf
\else
\fi
\usepackage{multirow}
\usepackage{bbding}
\usepackage{amsmath}
\usepackage{bm}
\usepackage{amssymb}
\usepackage{subfigure}  
\usepackage{graphicx}
\usepackage{cite}
\usepackage{acronym}
\usepackage{algorithmic}
\usepackage{array}
\usepackage{url}

\begin{document}
\title{Learning Deep Interleaved Networks with Asymmetric Co-Attention for Image Restoration}
\author{Feng~Li, Runmin~Cong, Huihui~Bai, Yifan~He, Yao~Zhao,~\IEEEmembership{Senior Member,~IEEE}, and Ce Zhu, ~\IEEEmembership{Fellow,~IEEE}
\IEEEcompsocitemizethanks{\IEEEcompsocthanksitem Feng Li, Yifan He, Runmin Cong, Huihui Bai, and Yao Zhao are with the Beijing Key Laboratory of Advanced Information Science and Network Technology, Beijing 100044, China; and also with the Institute of Information Science, Beijing Jiaotong University, Beijing 100044, China. Email: \{l1feng, rmcong, yifanhe, hhbai, yzhao\}@bjtu.edu.cn.
\IEEEcompsocthanksitem C. Zhu is with the School of Electronic Engineering, University of Electronic Science and Technology of China, Chengdu 611731, China, and also with the Center for Robotics, University of Electronic Science and Technology of China, Chengdu 611731, China (e-mail: eczhu@uestc.edu.cn)}
\thanks{This work was supported in part by the Fundamental Research Funds for the Central Universities (2019JBZ102) and the National Natural Science Foundation of China (No. 61972023). Corresponding author: Huihui Bai.}}

% The paper headers
\markboth{Journal of \LaTeX\ Class Files,~Vol.~14, No.~8, August~2015}%
{Shell \MakeLowercase{\textit{et al.}}: Bare Advanced Demo of IEEEtran.cls for IEEE Computer Society Journals}

\IEEEtitleabstractindextext{%
\begin{abstract}

Recently, convolutional neural network (CNN) has demonstrated significant success for image restoration (IR) tasks (e.g., image super-resolution, image deblurring, rain streak removal, and dehazing). However, existing CNN based models are commonly implemented as a single-path stream to enrich feature representations from low-quality (LQ) input space for final predictions, which fail to fully incorporate preceding low-level contexts into later high-level features within networks, thereby producing inferior results. In this paper, we present a deep interleaved network (DIN) that learns how information at different states should be combined for high-quality (HQ) images reconstruction. The proposed DIN follows a multi-path and multi-branch pattern allowing multiple interconnected branches to interleave and fuse at different states. In this way, the shallow information can guide deep representative features prediction to enhance the feature expression ability. Furthermore, we propose asymmetric co-attention (AsyCA) which is attached at each interleaved node to model the feature dependencies. Such AsyCA can not only adaptively emphasize the informative features from different states, but also improves the discriminative ability of networks. Our presented DIN can be trained end-to-end and applied to various IR tasks. Comprehensive evaluations on public benchmarks and real-world datasets demonstrate that the proposed DIN perform favorably against the state-of-the-art methods quantitatively and qualitatively. Code is available at \url{https://github.com/lifengshiwo/DIN}.

\end{abstract}

% Note that keywords are not normally used for peerreview papers.
}

% make the title area
\maketitle

\IEEEdisplaynontitleabstractindextext

\IEEEpeerreviewmaketitle

\ifCLASSOPTIONcompsoc
\IEEEraisesectionheading{\section{Introduction}\label{sec:introduction}}
\else
\section{Introduction}
\label{sec:introduction}
\fi

\IEEEPARstart{I}{mage} Restoration (IR), with the goal of recovering a high-quality (HQ) image from a degraded low-quality (LQ) observation, is a classical low-level vision problem and has drawn much attention in a variety of  applications. However, IR is an ill-posed inverse problem since there are multitudes of latent clean images can be corrupted to LQ one. In general, the degradation process of a HQ image $x$ can be formulated as $y = \mathcal{D}x + n$, where $\mathcal{D}$ denotes degradation operator, and $n$ denotes additive noise. $x$ and $y$ are the original HQ image and the corrupted counterpart, respectively. With different mathematical settings of $\mathcal{D}$ and noise type, one can correspondingly express different IR tasks. For example, IR task can be denoising problem~\cite{bm3d,nonlocalsparse,dncnn,denoiseprior} when $\mathcal{D}$ is defined as an identical matrix, deblurring problem~\cite{bayesian,idbm3d,scgmm} when $\mathcal{D}$ is a blur kernel, and super-resolution (SR)~\cite{srcnn,selfexsr,vdsr,edsr} when $\mathcal{D}$ is a composite operator of a blur convolution and a subsampling matrix. 

To tackle these IR problems, in the past few decades, numerous algorithms have been proposed to reconstruct the underlying HQ image from the single LQ input including model-based methods~\cite{denoiseprior,scgmm,sparsedl,nonlocalir} and learning-based methods~\cite{dncnn,ircnn,memnet,msdeblur,ddn}. Model-based methods consider IR as an optimization problem based on the maximum a posterior (MAP) framework, which design image prior as regularization term, such as total variation (TV)~\cite{tv}, sparsity~\cite{sparsedl,sparseir}, and non-local means~\cite{bm3d,nonlocalsparse,nonlocalir}, \emph{et al.}, to enforce desired property of HQ output. Learning-based methods aims at learning mapping functions to restore desirable HQ images from the distorted input. Inspired by the powerful leaning ability of convolutional neural networks (CNN) in high-level vision problems~\cite{vgg,resnet,densenet}, many CNN models have been proposed for a variety of IR tasks, such as image super-resolution (SR)~\cite{srcnn,vdsr,edsr,srresnet,rdn}, motion deblurring~\cite{msdeblur,deblurgan}, deraining~\cite{ddn,djr}, and image dehazing~\cite{dehazenet,gfn}. Dong~\emph{et al.}~\cite{srcnn} first employ shallow CNN for image SR and significantly outperform conventional algorithms. VDSR~\cite{vdsr} and DnCNN~\cite{dncnn} increase the depth to exploit the contextual information over large regions and employ global residual learning to ease training difficulty. Lim~\emph{et al.}~\cite{edsr} modify the residual blocks in SRResNet~\cite{srresnet} and employ a much deeper and wider network for image SR (EDSR). 

%%%%
\begin{figure*}
\centering
\includegraphics[width=7.12in]{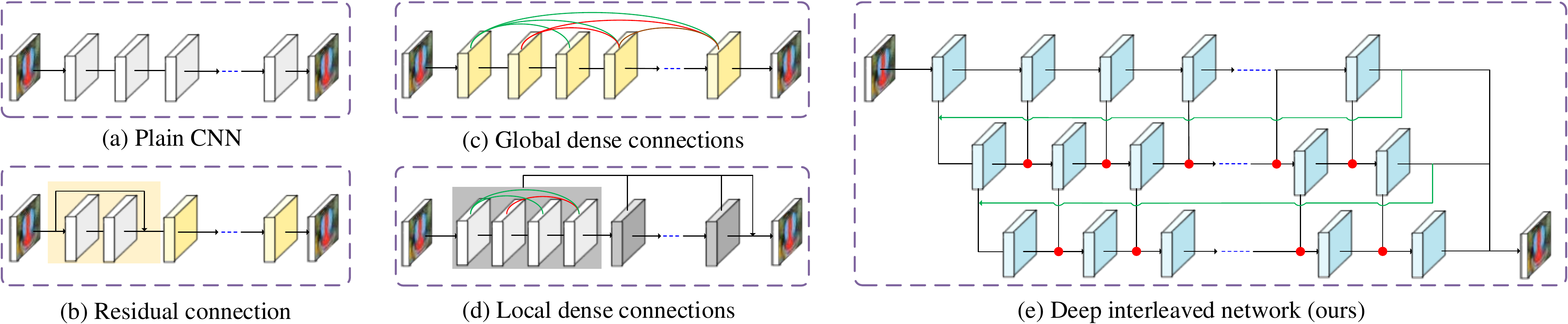}
\caption{ Comparisons of Deep IR Networks. (a) Plain CNN (\emph{e.g.}, SRCNN~\cite{srcnn}, VDSR~\cite{vdsr} and DnCNN~\cite{dncnn}). (b) Residual connections (\emph{e.g.}, EDSR~\cite{edsr} and SRResNet~\cite{srresnet}). (c) Global dense connections (\emph{e.g.}, MemNet~\cite{memnet} and CSFM~\cite{csfm}). (d) Local dense connections (\emph{e.g.}, SRDenseNet~\cite{srdensenet} and RDN~\cite{rdn}). (e) The basic structure of our proposed DIN, which is a multi-path and multi-branch framework. The light blue block is our built multi-path block in DIN, which will be shown in Section~\ref{sec:wrdb}.}
\label{fig0}
\end{figure*}
%%%%

Hierarchical features under different receptive fields can provide abundant information. It is crucial to incorporate features from multiple layers for IR tasks. Some methods~\cite{vdsr,edsr,dncnn,lapsrn,srresnet} directly stack convolutional layers~\cite{vdsr,dncnn,lapsrn} or build ResNet-like blocks~\cite{edsr,srresnet}, which fail to fully exploit hierarchical features and thus cause relatively-low SR performance. Existing methods adopt two ways for this problem: 1) \emph{Global dense connections.} Tai~\emph{et al.}~\cite{memnet} present to explicitly mine persistent memory through densely connected memory blocks, which can help information flow between intermediate states for IR problems. Hu~\emph{et al.}~\cite{csfm} utilize gated fusion nodes to concatenate hierarchical features from preceding modules. However, inside of the internal blocks of these methods, the subsequent convolutional layers don't have access to receive information from preceding layers. 2) \emph{Local dense connections and global fusion}. Tong \emph{et al.}~\cite{srdensenet} leverage the basic block in DenseNet~\cite{densenet} to obtain abundant dense features from different levels to boost the reconstruction performance. Zhang~\emph{et al.}~\cite{rdn} propose a residual dense network (RDN) to extract abundant local features with the combination of dense connections and residual learning. Nevertheless, the two methods only utilize dense connections to fuse hierarchical features from all the convolutional layers within each block, the information from these blocks are only combined at the tail of networks. By this way, preceding low-level features lack full incorporation with later high-level features. These basic design methodologies are illustrated in Fig.~\ref{fig0}.

In this work,  to address this problem, we propose a novel deep interleaved network (DIN) that aggregates multi-level features and learns how information at different states should be combined for IR tasks. We enrich the feature representations from two aspects: internal multi-path blocks and cross-branch connections. Specifically, we propose weighted residual dense block (WRDB) to exploit hierarchical features that gives more clues for better reconstruction. The key component of WRDB lies in the densely weighted connections (DWC) among several residual dense blocks (RDB). In the WRDB, besides the direct connection from current RDB to the next one, we assign different weighted parameters to different outputs from preceding states fed into all the subsequent RDBs. The internal multi-path RDBs and DWCs can ensure persistent memory and more precise features aggregation. Then, to mitigate the restricted multi-level context incorporation of solely feed-forward architectures, we implement an interconnected cross-branch framework, where each branch consists of multiple cascading WRDBs.  Between adjacent branches, the WRDBs are interleaved horizontally and vertically to progressively incorporate multi-level contexts from different states. With this methodology, the shallow information guide deep representative features prediction, and later branches can generate more powerful feature representations in combination with former branches.

Besides, at each interleaved node between adjacent branches, we consider to implement an effective method to adjust received information from different states for feature fusion. The features from different states have different degrees of importance for the representational ability of networks. Commonly element-wise summation and concatenation treat all input features equally, which lack flexibility to modulate these features. To handle this issue, in this paper, we propose asymmetric co-attention (AsyCA) that performs feature recalibration and learns to adaptively emphasize input informative features. The AsyCA are attached at the interleaved nodes for trainable weights generation to fuse important information from different states. Hence, the AsyCA can effectively improve the discriminative capacity of our DIN for high-frequency details. Moreover, global feature fusion (GFF) and residual learning (GFL) are utilized to further boost the restoration performance.  

Overall, the main contributions of our work are summarized as follows:
\begin{enumerate}
\item We propose a novel deep interleaved network (DIN) which employs a multi-path and multi-branch framework to fully exploit informative hierarchical features and learn how information at different states should be combined for various IR tasks. 
\item We propose weighted residual dense block (WRDB) composed of multiple residual dense blocks in which different weighted parameters are assigned to different inputs for more precise features aggregation and propagation. 
\item In the proposed DIN, asymmetric co-attention (AsyCA) is proposed and attached to the interleaved nodes to adaptively emphasize informative features from different states, which can effectively improve the discriminative ability of networks for high-frequency details recovery.
\item Extensive experiments results on image super-resolution, motion blur removal, deraining, and dehazing have demonstrated the superiority of our DIN over state-of-the-art methods.
\end{enumerate}

A preliminary version of this work has been presented as a conference version earlier~\cite{din}. In this work, we add to the initial version in significant ways: 1). According to the characteristics of different IR tasks, we research a flexible framework of DIN to meet the need of different problems. 2) Considerable new analyses and intuitive ablations are added to the initial results. We also extend the original image SR experiments from $2\times$, $3\times$, $4\times$ to $8\times$ and other degradation models, such as blur degradation (\textbf{BD}) and noise degradation (\textbf{DN}). 3). We extend the DIN to process a variety of IR tasks including image SR, motion deblurring, rain streak removal and image dehazing. It is shown that the proposed DIN achieves superior performance in comparison with existing methods for these tasks, respectively. 

\section{Related Work}
Numerous IR methods have been extensively studied in the literature. In this section, we briefly review several IR tasks that are related to the proposed method.

\subsection{Image Super-Resolution}
Image SR has gained increasing attention and achieved dramatic improvements using deep learning based methods in recent years. As a pioneer work, Dong~\emph{et al.}~\cite{srcnn} propose SRCNN to minimize the mean square error (MSE) between the bicubic-interpolated image and HR image. Kim~\emph{et al.}~\cite{vdsr} construct a very deep SR network (VDSR) with global residual learning that improves the reconstruction performance with more convolutional layers. DRCN~\cite{drcn} introduce recursive supervision that shares weights across different layers to reduce the model complexity. Tai~\emph{et al.}~\cite{drrn} combines global/local residual connections and recursive learning to control the number of parameters while increasing the depth. In recent years, many methods~\cite{lapsrn,edsr,rdn,espcn,srfbn} take original LR images as input and leverage transposed convolution~\cite{lapsrn,srfbn} or sub-pixel layer~\cite{espcn} to upscale final learned LR feature maps into HR space. Lai~\emph{et al.}~\cite{lapsrn} propose LapSRN to progressively reconstruct the sub-band residuals of HR images at multiple pyramid levels. Haris~\emph{et al.}~\cite{dbpn} introduce DBPN which incorporates error feedbacks in multiple iterative up- and down stages to accumulate the self-correcting features for image SR. Li~\emph{et al.}~\cite{srfbn} present SRFBN that combines recurrent neural network (RNN) and feedback mechanism to refine low-level representations with high-level information. Motivated by the efficiency of the sub-pixel convolution in ESPCN~\cite{espcn}, some models \cite{edsr,rdn,san} adopt the similar post-upsampling approach at the end of networks to recover HR images.

\subsection{Motion Deblurring}
Earlier works~\cite{xu,learndeblur,blinddeblur} mainly remove motion blurs depending on synthesized images with uniform blur kernel, which cannot suit spatially vary blurs. Recently, Gong~\emph{et al.}~\cite{heterogeneous} propose to estimate pixel-wise motion flow and remove heterogeneous motion blue using by learning from simulated examples. In~\cite{msdeblur}, a multi-scale CNN is presented to remove non-uniform blind blurs in a coarse-to-fine manner. Zhang~\emph{et al.}~\cite{zhang} construct a recurrent encoder-decoder network and introduce temporal feature transfer for motion deblurring. Kupun~\emph{et al.}~\cite{deblurgan} introduce DeblurGAN based on conditional generative adversarial network (GAN)~\cite{cgan} to handle blind motion deblurring by optimize a multi-component loss function upon a hand-crafted datasets with more complex blur kernels.

\subsection{Rain Streak Removal}
Rain streaks in images and videos often cause severely undesirable impact on outdoor vision conditions. Traditional optimization based methods~\cite{automatic,gmm} separate the rain streaks based on the statistical properties. In the past few years, we have witnessed unprecedented progress of CNN in image deraining. Fu~\emph{et al.}~\cite{ddn} propose deep detail network (DDN) which decomposes the rainy image into background and high-frequency rain content by a detail layer for image deraining. Yang~\emph{et al.}~\cite{djr} introduce region-dependent models to jointly detect and remove raindrops via a multi-task learning architecture. In~\cite{rescan}, RESCAN divides the rain removal into multi-stages and uses contextual aggregation network in each stage to remove the rain streaks. Ren~\emph{et al.}~\cite{prenet} propose PReNet which combines multi-stage recursion and conventional ResNet to progressively remove the rain streaks. In~\cite{spanet}, Wang~\emph{et al.} incorporate temporal priors and human supervision to remove the rains from observed rainy image sequences. 

\subsection{Image Dehazing}
Haze is an atmospheric phenomenon where dust, smoke, and other obscure particulates cause blurring and contrast loss for the photographed image. He~\emph{et al.}~\cite{dark} investigate the statistics of haze-free outdoor images and propose a dark channel prior based algorithm. Meng~\emph{et al.}~\cite{boundary} combine the inherent boundary constraint on the transmission function with weighted $L_1$-norm based contextual regularization for haze removal. Recently, CNNs also have been utilized to tackle this problem~\cite{dehazenet,gfn,mscnn,dcpdn,aodnet}. DehazeNet~\cite{dehazenet} conducts medium transmission estimation to recover a clean image via an atmospheric scattering model. Ren~\emph{et al.}~\cite{gfn} construct GFN that computes pixel-wise confidence map based on white balance, contrast enhancing, and gamma correction for image dehazing. In~\cite{dcpdn}, Zhang~\emph{et al.} adopt multi-level pyramid pooling module to estimate transmission map and introduce an edge-preserving loss to enhance dehazed results. Qu~\emph{et al.}~\cite{epdn} propose enhanced pix2pix dehazing network that embeds an enhancer to reinforce the output of the GAN for haze removal.

\section{Proposed Algorithm}
We take various image restoration problems into consideration, where the degradation patterns of these problems are known and the paired images of corrupted input and ground truth are available. In this section, we first elaborate the core components of our deep interleaved network (DIN) in detail including the weighted residual dense block (WRDB), asymmetric co-attention (AsyCA), and interleaved multi-branch framework (IMBF). Then, we introduce the architecture of  DIN and provide the analysis of specific cases for different tasks.

%%%%
\begin{figure}[h]
\centering
	  \subfigure[Common residual dense block (RDB)]{\label{fig1:subfig:a} 
       \includegraphics[width=3.4in]{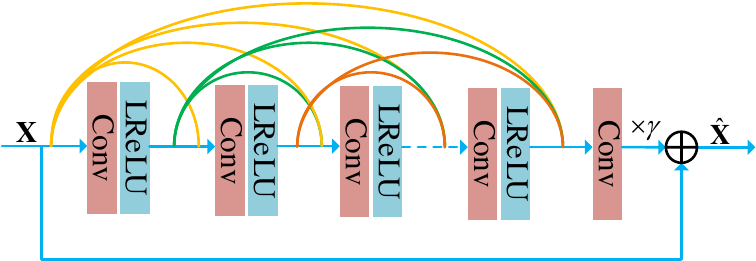}}
		\hfill
	  \subfigure[Weighted residual dense block]{\label{fig1:subfig:b}
        \includegraphics[width=3.4in]{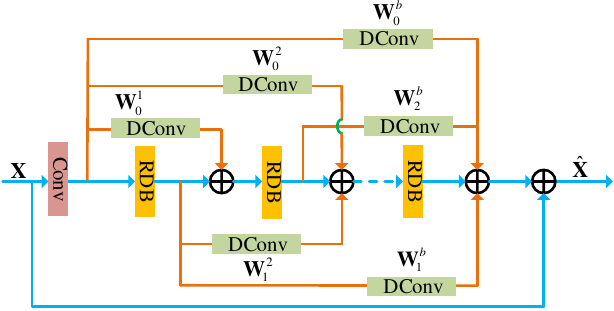}}
		\hfill
	  \caption{The structures of (a) common residual dense block (RDB) and (b) the proposed weighted residual dense block (WRDB).}
	  \label{fig1}
\end{figure}
%%%%

\subsection{Weighed Residual Dense Block} 
\label{sec:wrdb}
In order to exploit hierarchical features that enhance the quality of reconstructed HR images, we present weighted residual dense block (WRDB) which utilizes weighted identity shortcuts to combine immediate states and propagate more precise information flow across these states. Compared to common residual dense block (RDB, Fig. \ref{fig1:subfig:a}), the main differences between WRDB and RDB lie in the densely weighted connections (DWCs) and local cascading construction, where Fig. \ref{fig1:subfig:b} depicts the structure of WRDB.

\subsubsection{Revisiting RDB}
Firstly, supposing there are $K$ layers in one RDB, later layers receive the features from all preceding layers as input, the output of the $k^{th}$ convolutional layer is
%%%%
\begin{equation}
F^b_k = \sigma (w^b_k \ast [F^{b-1}, F^b_{1},  F^b_{2},..., F^b_{k-1}])
\label{eq1}
\end{equation}
%%%% 
where $w^b_k$ is the weights of the $k^{th}$ layer in the $b^{th}$ RDB and the bias term is omitted for simplicity. $[F^{b-1}, F^b_{1},  F^b_{2},..., F^b_{k-1}]$ represents the concatenation of
the feature maps generated in the preceding layers. $F^{b-1}$ is the output of last RDB and $\sigma(\cdot)$ denotes the LeakyReLU (LReLU)~\cite{lrelu} activation function. We also exploit residual learning between input feature $F^{b-1}$ and output feature $F^b_K$ to help the information flow 
%%%%
\begin{equation}
F^b = F^b_K + \gamma \ast F^{b-1}
\label{eq2}
\end{equation}
%%%%
where $F^b$ is the final output of the $b^{th}$ RDB and $\gamma$ is a constant as the residual scaling parameter to keep the training stability (Fig. \ref{fig1:subfig:a}). We set $\gamma=0.1$ in our experiments.

\subsubsection{Densely Weighted Connections}
According to the investigation in~\cite{densenet,srdensenet,rdn} that dense connections can reuse the feature maps from preceding layers and improve the information flow through networks, the proposed WRDB employs RDB as the basic element and builds a more effective structure. Specifically, as shown in Fig. \ref{fig1:subfig:b}, given an input feature $\mathbf{X}\in\mathbb{R}^{H\times W\times C}$ where $H$ and $W$ are the spatial height and width of $C$ feature maps, $B$ RDBs are utilized to extract local dense feature. Weighted connections are created between a RDB and every other RDB. In one WRDB, the output of $b^{th}$ RDB can be formulated as 
%%%%
\begin{equation}
\begin{aligned}
\mathbf{X}_b &= sum(\mathbf{W}^b_0 \ast \mathbf{X}_0, \mathbf{W}^b_1 \ast \mathbf{X}_1,..., \mathbf{W}^b_{b-1} \ast \mathbf{X}_{b-1})\\
\mathbf{X}_0 & = H_{0, d}(\mathbf{X})\\
\end{aligned}
\label{eq3}
\end{equation}
%%%% 
where $sum(\cdot)$ represents element-wise summation. $H_{0, d}(\cdot)$ denotes the convolution operation of the first convolutional layer in WRDB. $\mathbf{W}^b_0$ denotes the scaling weight set for the shortcut between the first convolutional layer and the $b^{th}$ RDB and $\mathbf{W}^b_{b-1}$ denotes the weight set that rescale the output feature of the $(b -1)^{th}$ block that are fed into the $b^{th}$ block. Therefore, all the features from preceding states are rescaled by our densely weighted connections (DWC) and then fused at current state. 

%%%%
\begin{figure}
\centering
\includegraphics[width=2.7in]{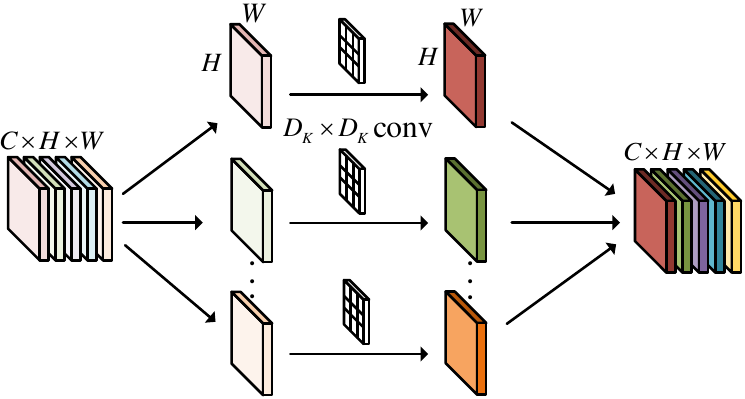}
\caption{The convolution processing of depth-wise convolutional filters.}
\label{fig2}
\end{figure}
%%%%

In contrast to simply using hyper-parameter or learned scalar for feature rescaling, we leverage depth-wise convolution to perform weight operation. As we know, each kernel in standard convolution filters all the input features and combines them to produce a new representation. Compared to standard convolution, as shown in Fig. \ref{fig2},  given an input feature $\mathbf{F}\in\mathbb{R}^{C\times H\times W}$ with $C$ channels, 
depth-wise convolution applies one kernel on each channel respectively, which only conducts filtering operation but not combine them. Thus, each kernel in depth-wise convolution can be regarded as a weight parameter of an individual feature map. Besides, the computational cost of depth-wise convolution can be calculated as
%%%%
\begin{equation}
D_K \times D_K \times C \times H \times W
\label{eq4}
\end{equation}
%%%%
where $D_K$ is the spatial dimension of the kernel. In this work, the kernel size is set as $1\times1$ which involves very low computation and parameters. Based on the above analysis, the depth-wise convolutional layers can be easily applied to the immediate states within our WRDB to balance the importance of all preceding features and pass more detailed information flow across multiple states. Finally, the output of the $d^{th}$ WRDB can be obtained via local residual learning
%%%%
\begin{equation}
\hat{\mathbf{X}} = \mathbf{X} + \mathbf{X}_B
\label{eq5}
\end{equation}
%%%%
where $\mathbf{X}_B$ denotes the output of the $B^{th}$ RDB in the WRDB.

\subsection{Asymmetric Co-Attention}
\begin{figure}
\centering
\includegraphics[width=3.36in]{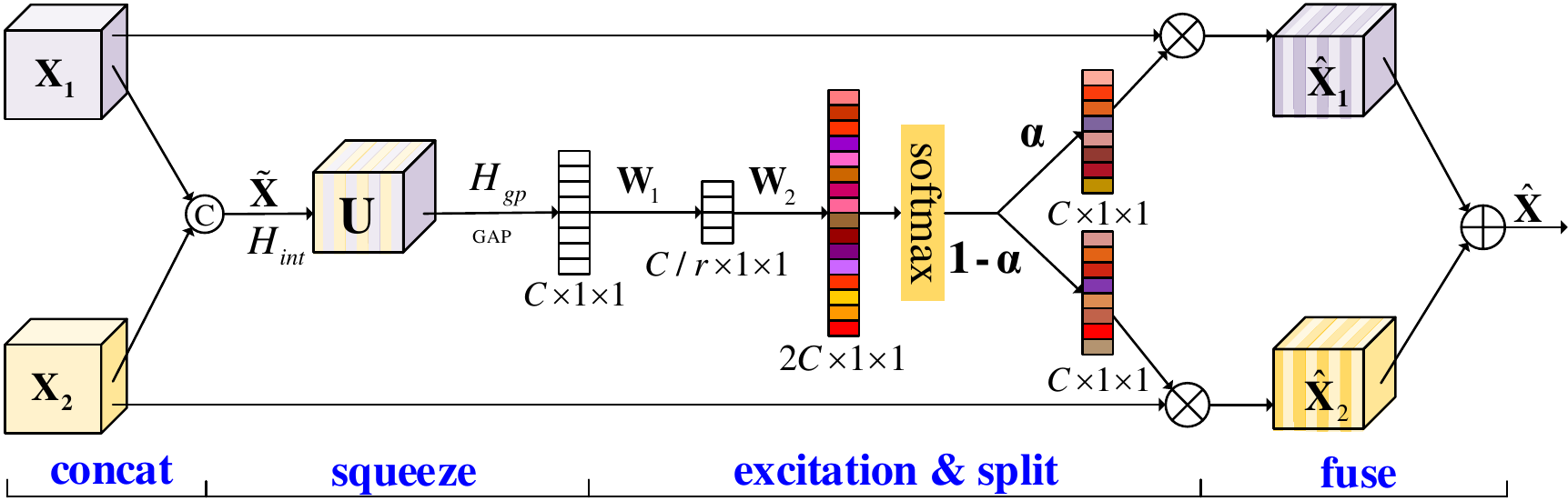}
\caption{The proposed asymmetric co-attention (AsyCA) architecture. ``GAP'' denotes the global average pooling.}
\label{fig3}
\end{figure}

In a deep network, the features from different states make different contributions to the learning of latter feature representations. Simply combining these features using local residual connections or concatenation lack flexibility to modulate these features, which resultantly limits the discriminative ability of deep models. The proposed asymmetric co-attention (AsyCA) aims at adaptively emphasizing important information from different states. The structure of our AsyCA is sketched in Fig. \ref{fig3}, which consists of four steps: concat, squeeze, excitation \& split, and fuse. As we can see,  AsyCA receives features from multiple states and generates trainable weights for feature fusion. Here, we use the features from two states to depict the details. 

%%%%
%%%%

\textbf{Concat}. Given features $\mathbf{X}_1$ and $\mathbf{X}_2$ are both with size of $C\times H\times W$, we first conduct concatenation operation on the two features
%%%%
\begin{equation}
\tilde{\mathbf{X}} = concat(\mathbf{X}_1, \mathbf{X}_2)
\label{eq6}
\end{equation}
%%%%
where $concat(\cdot)$ denotes the concatenation operation. Then, we adjust and integrate the coming information flow, which are performed by one $1\times1$ convolutional layer
%%%%
\begin{equation}
\mathbf{U} = H_{int}(\tilde{\mathbf{X}})
\label{eq7}
\end{equation}
%%%%
where $H_{int}(\cdot)$ indicates the integration function that has $C$ filters. Hence we can write the output as $\mathbf{U}=[\mathbf{u}_1, \mathbf{u}_2,..., \mathbf{u}_C]$ that consists of $C$ feature maps with the size of $H\times W$.

\textbf{Squeeze}. To ensure our network are sensitive to informative features and model the channel interdependencies for feature recalibration, we would like to squeeze the global spatial information of $\mathbf{U}$ into a channel descriptor. To this end, global average pooling is utilized to produce a channel-wise summary statistics $\mathbf{z}\in\mathbb{R}^{C\times1\times1}$.  The $c^{th}$ element of $\mathbf{z}$ can be computed by shrinking $\mathbf{U}$ through spatial dimensions $H\times W$
%%%%
\begin{equation}
z_c = H_{gp}(\mathbf{u}_c) = \frac{1}{H\times W}\sum^H_{i=1}\sum^{W}_{j=1}\mathbf{u}_c(i, j)
\label{eq8}
\end{equation}
%%%%
where $\mathbf{u}_c(i, j)$ is the value at position $(i, j)$ of the $c^{th}$ channel $\mathbf{u}_c$. $H_{gp}(\cdot)$ denotes global average pooling operation.

\textbf{Excitation \& split}. The goal of excitation operation is to capture the channel-wise dependencies based on the squeezed channel-wise statistics $\mathbf{z}$. Here, two $1\times1$ convolutional layers are used to form a bottleneck that performs dimensionality-reduction and -increasing with reduction ratio $r$.
%%%%
\begin{equation}
\mathbf{s} = \mathbf{W}_2\ast\delta(\mathbf{W}_1\ast\mathbf{z})
\label{eq9}
\end{equation}
%%%%
where $*$ denotes convolution operation and $\delta(\cdot)$ represents the ReLU~\cite{relu} activation function. $\mathbf{W}_1\in\mathbb{R}^{C/r\times C\times H\times W}$ and $\mathbf{W}_2\in\mathbb{R}^{2C\times C/r\times H\times W}$ are the learned weights of the two convolutional layers. We set $r =4$ in all of our experiments. In this way, we can obtain the channel statistics $\mathbf{s}$ with the size of $2C\times1\times1$ that are corresponding to the received information flows. 

Next, for the input $\mathbf{X}_1$ and $\mathbf{X}_2$, there are totally $2C$ feature maps, where each input has $C$. We need to map the feature descriptor $\mathbf{s}$ to a set of channel-specific weights and adaptively emphasize the two information flows. For this purpose, we employ a softmax operator to calculate the attention weighs across channels and split the output into two chunks
%%%%
\begin{equation}
\begin{aligned}
\alpha_c &= \frac{exp(\mathbf{V}^c_1\mathbf{s})}{exp(\mathbf{V}^c_1\mathbf{s})+exp(\mathbf{V}^c_2\mathbf{s})}\\
1-\alpha_c &= \frac{exp(\mathbf{V}^c_2\mathbf{s})}{exp(\mathbf{V}^c_1\mathbf{s})+exp(\mathbf{V}^c_2\mathbf{s})}
\end{aligned}
\label{eq10}
\end{equation}
%%%%
where $\mathbf{V}_1\in\mathbb{R}^{C \times 1\times 1}$ and $\mathbf{V}_2\in\mathbb{R}^{C \times 1\times 1}$ denote the attention vector of $\mathbf{X}_1$ and $\mathbf{X}_2$, respectively. $\mathbf{V}^c_1$ is the $c^{th}$ row of  $\mathbf{V}_1$ and $\alpha_c$ is the corresponding element of $\bm{\alpha}$.

\textbf{Fuse}. We obtain the attentive feature $\widehat{\mathbf{X}}$ by conducting channel-wise multiplication between rescaling the input features $\mathbf{X}_1$ and $\mathbf{X}_2$ and resulted attention weights
%%%%
\begin{equation}
\widehat{\mathbf{x}}_c = \alpha_c\mathbf{x}_{1, c} + (1 - \alpha_c)\mathbf{x}_{2, c}
\label{eq11}
\end{equation}
%%%%
where $\mathbf{x}_{1,c}$, $\mathbf{x}_{2,c}$, and $\widehat{\mathbf{x}}_c\in\mathbb{R}^{H\times W}$ denote the $c^{th}$ elements of $\mathbf{X}_1$, $\mathbf{X}_2$, and $\widehat{\mathbf{X}}$, respectively. Therefore, the output of our AsyCA can be represented as
%%%%
\begin{equation}
\widehat{\mathbf{X}} = \pmb{\alpha} \cdot \mathbf{X}_1 + (\mathbf{1} - \pmb{\alpha}) \cdot \mathbf{X}_2
\label{eq12}
\end{equation}
%%%%
where $\widehat{\mathbf{X}}=[\widehat{\mathbf{x}}_1, \widehat{\mathbf{x}}_2, ..., \widehat{\mathbf{x}}_c]$. 

Conclusively, the proposed AsyCA transforms input features into compact descriptor and generate two sets of channel-specific weights to model channel-wise interdependencies. In this way, AsyCA can adaptively emphasize the important information from various states and generate trainable weights for feature fusion, which is beneficial for discriminative learning. It should be noticed that AsyCA can be extended to more states than the two shown in Fig. \ref{fig3} and easily inserted in deep networks.

%%%%
\begin{figure*}[t]
\centering
\includegraphics[width=6.4in]{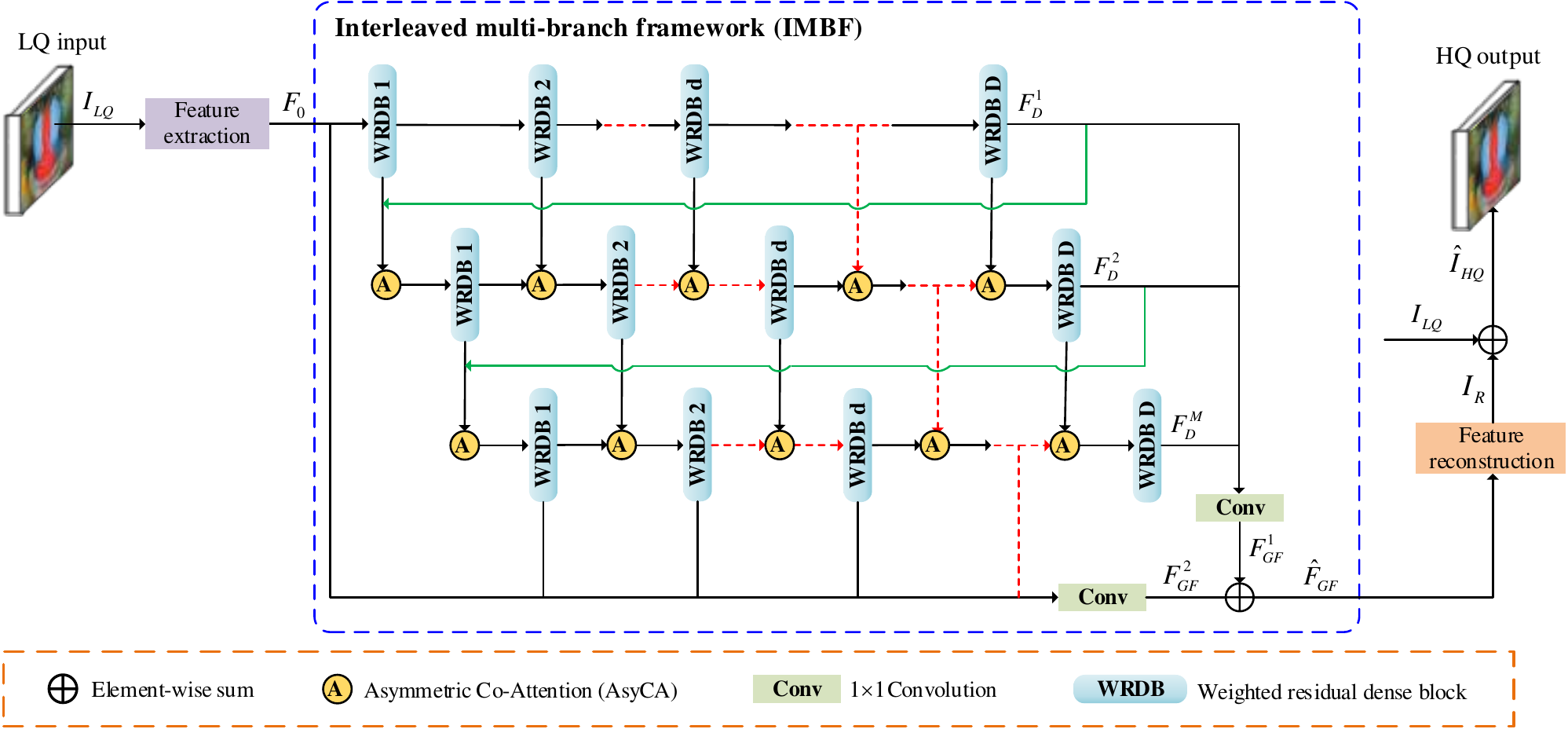}
\caption{The structural implementation of the proposed DIN for various IR tasks, where the input can be LR, blurry, rainy, or hazy images. Here, we only use LQ input for simply description.}
\label{fig4}
\end{figure*}
%%%%

\subsection{Interleaved Multi-Branch Framework}
Our DIN follows a multi-branch pattern allowing multiple interconnected branches to interleave and fuse at different states. Now we present the backbone, interleaved multi-branch framework (IMBF), in DIN for IR. As illustrated in Fig. \ref{fig4}, the main point of IMBF lies in a progressively cross-branch feature interleaving, and cascading WRDBs in each branch. We define the first branch as backbone which consists of $D$ WRDBs. Given an input feature $F_0$, the output of this basic branch can be represented as
%%%%
\begin{equation}
F^1_D = H^1_D(H^1_{D-1}(...(H^1_1(F_0))...))
\label{eq13}
\end{equation}
%%%%
where $H^1_D(\cdot)$ denotes the operation of the $D^{th}$ WRDB in the first branch, which can be a composite function of operations. $F^1_D$ is the output feature. Then, we iteratively replicate the backbone many times to form a multi-branch framework. Therefore, in IMBF, each branch has the same structure but dose not share weight parameters. 

The sub-network in each branch can be regarded as a refinement process by continuously fusing the features from different states. Specifically, as shown in Fig. \ref{fig4}, for two adjacent branches, the features in the same depth from these branches respectively are combined to incorporate preceding low-level contexts into current information flow. The output of a certain block in the previous branch is contributed to the input of its corresponding block in current branch. For the $m^{th}$ branch, the learning process of the $d^{th}$ block can be defined as $H^m_d(\cdot)$. Our
multi-branch feature interleaving can be formulated as
%%%%
\begin{equation}
F^m_d=\left\{
             \begin{array}{lr}
             H^m_d(S^m_d([F^{m-1}_D, F^{m-1}_1])), & \text{if}\; d=1 \\
             H^m_d(S^m_d([F^{m-1}_d, F^m_{d-1}])), & otherwise.\\
             \end{array}
\right.
\label{eq14}
\end{equation}
%%%%
where $S^m_d (\cdot)$ denotes the fusion operation at the interleaved nodes (orange circles in Fig. \ref{fig4}). $F^m_d$ is the output of the $d^{th}$ block in branch $m$. $m$ and $d$ are integers in range $[1, M]$ and $[1, D]$, respectively. We can observe that, at the initial state in every branch, the first and last WRDBs in the previous branch are fused as input fed into current branch. 

Finally, global feature fusion (GFF) strategy is adopted to aggregate features from all branches 
%%%%
\begin{equation}
\begin{aligned}
F^1_{GF} &= H^1_{GFF}(F^1_D, F^2_D, ..., F^{M-1}_D, F^M_D) \\
F^2_{GF} &= H^2_{GFF}(F^M_1, F^M_2, ..., F^M_{D-1}, F_0) \\
\end{aligned}
\label{eq15}
\end{equation}
%%%%
where $H^1_{GFF}(\cdot)$ and $H^2_{GFF}(\cdot)$ denote the convolution operations of the two convolutional layers at the tail of IMBF to integrate received deep features, where each layer has 64 filters with the kernel size of $1\times1$. After that, element-wise addition is used to combine the two integrated features
%%%%
\begin{equation}
\hat{F}_{GF} = F^1_{GF} + F^2_{GF} 
\label{eq16}
\end{equation}
%%%% 
where $\hat{F}_{GF}$ is the output of our IMBF.

Benefiting from this structure, the network can progressively combine the contextual information from different states to generate more powerful feature representations. It should be noticed that every feature fusion operation can be used at the interleaved nodes, such as element-wise addition, concatenation, or even SE block~\cite{senet}. In this work, we use the proposed AsyCA to make more representative feature fusion. More discussion will be shown in Section~\ref{sec:ablation}.

\subsection{Network Architecture for IR}
As can be seen in Fig.~\ref{fig4}, the whole architecture of our DIN can be divided into three parts: feature extraction, interleaved multi-branch framework (IMBF), and feature reconstruction. 

Here, let's denote $I_{LQ}$ as the LQ input image of DIN and $I_{HQ}$ is the corresponding ground-truth HQ image. In DIN, a feature extraction module is first employed to extract the shallow feature from $I_{LQ}$
%%%%
\begin{equation}
F_0 = H_0(I_{LQ})
\label{eq17}
\end{equation}
%%%%
where $H_0(\cdot)$ denotes convolution operation. $F_0$ is the extracted feature and served as an input that is fed into further modules for deep feature extraction. Then, we utilize the proposed IMBF to perform hierarchical feature interleaving and fusion
%%%%
\begin{equation}
\hat{F}_{GF} = H_{IMBF}(F_0)
\label{eq17}
\end{equation}
where $H_{IMBF}(\cdot)$ denotes the function of IMBF to produce a comprehensive feature $\hat{F}_{GF}$ (same as Eq.~(\ref{eq16})). Next, a feature reconstruction module and global residual learning (GRL) are used to generate HQ image
%%%%
\begin{equation}
\begin{aligned}
\hat{I}_{HQ} &= H_{FR}(\hat{F}_{GF}) + I_{LQ}\\ 
&= I_R + I_{LQ}
\end{aligned}
\label{eq18}
\end{equation}
%%%%
where $H_{FR}$ denotes the feature reconstruction function to produce $I_R$. Noticed that, in image SR, we use the bicubic interpolated HR image for GRL.

We use $L_1$ loss to optimize our DIN. Given a training dataset with $N$ LQ images and their HQ counterparts, denoted as $\{I^i_{LQ}, I^i_{HQ}\}^N_{i=1}$, the goal of training DIN is to optimize the $L_1$ loss function
%%%%
\begin{equation}
\begin{aligned}
L(\Theta) &= \frac{1}{N}\sum^N_{i=1}\Vert H_{DIN}(I^i_{LQ}; \Theta) - I^i_{HQ}\Vert_1\\
&= \frac{1}{N}\sum^N_{i=1}\Vert\hat{I}^i_{HQ} - I^i_{HQ}\Vert_1
\end{aligned}
\label{19}
\end{equation}
%%%%
where $\Theta$ denotes the learned parameter set of DIN and $H_{DIN}(\cdot)$ is the LQ-to-HQ mapping function.

\subsection{Implementation Details}
In our proposed DIN, we set the number of branches $M$ as 4, where each branch contains $D$ WRDBs. All the WRDBs are composed of one $3\times3$ convolutional layer followed by $B$ RDBs. The number of convolutional layers per RDB is 6, and the growth rate is 32. The convolutional layer in each RDB has 64 filters followed by LeakyReLU~\cite{lrelu} with negative slope value 0.2. We utilize $1\times1$ depth-wise convoluion to conduct densely weighted connections (DWCs) in every WRDB. The feature extraction module for motion deblurring and dehazing tasks contains one 64-channel convolutional layer with the size of $3\times3$ and one strided layer with size of $8\times8$ for $4\times$ downsampling. As for image SR and deraining, the feature extraction module only has one $3\times3$ convolutional layer. Correspondingly, in feature reconstruction module, for deblurring and dehazing, we use $4\times$ sub-pixel layer in ~\cite{din} for feature upscaling and HQ image reconstruction. This part for deraining only use single $3\times3$ convolutional layer to reconstruct HQ images. We adopt sub-pixel layer in \cite{din} as our feature reconstruction module to super-resolve LR features to HR image for $2\times$, $3\times$, $4\times$, and $8\times$ SR.
%%%%

\section{Ablation Study}
\label{sec:ablation}
In this section, we investigate the network structure and each key components in DIN. All experiments are conducted on image SR task for simplicity. 

%%%%
\begin{figure}[t]
\centering
	  \subfigure[]{\label{fig6:subfig:a} 
       \includegraphics[width=1.69in]{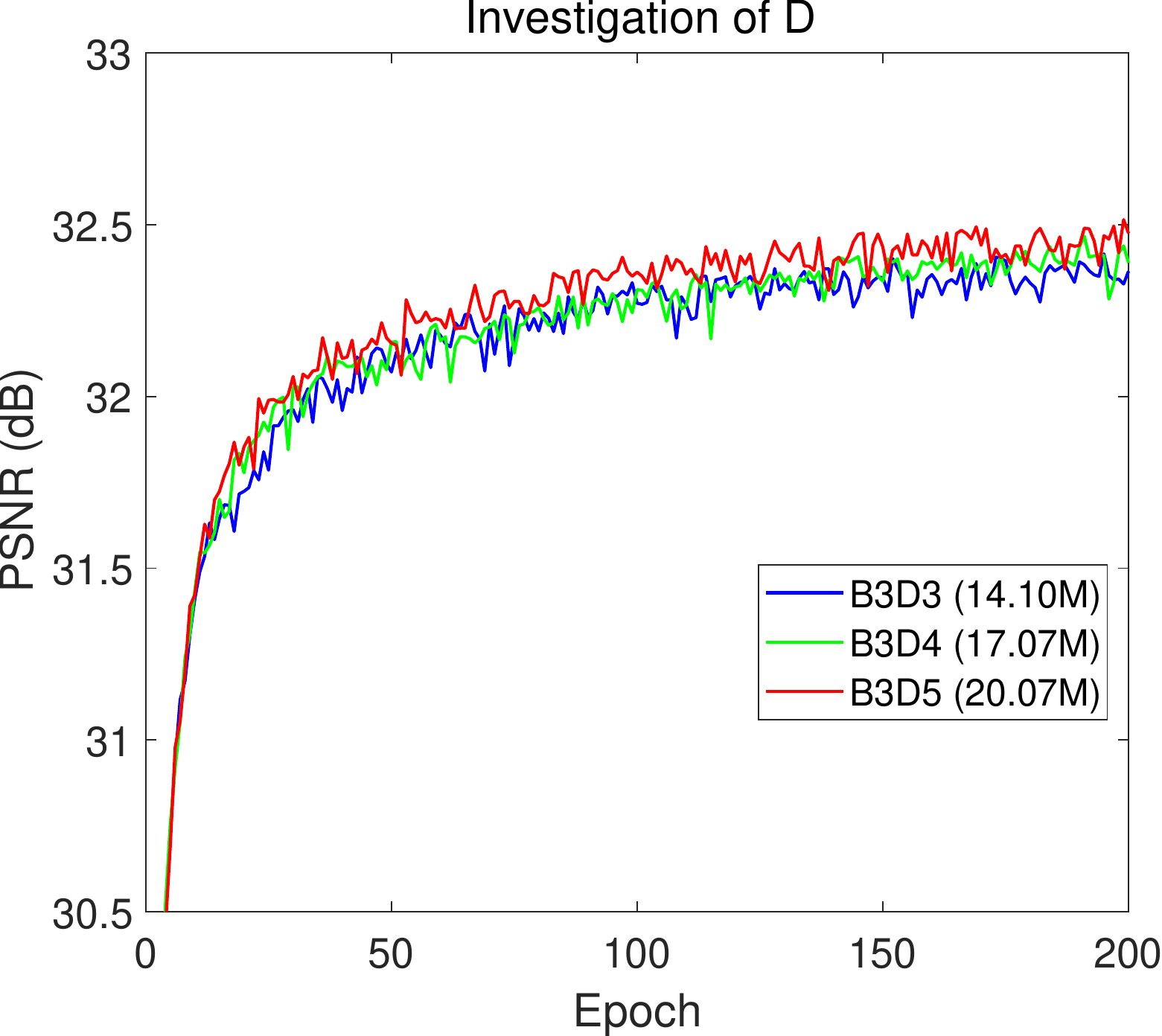}}
		\hfill
	  \subfigure[]{\label{fig6:subfig:b}
        \includegraphics[width=1.69in]{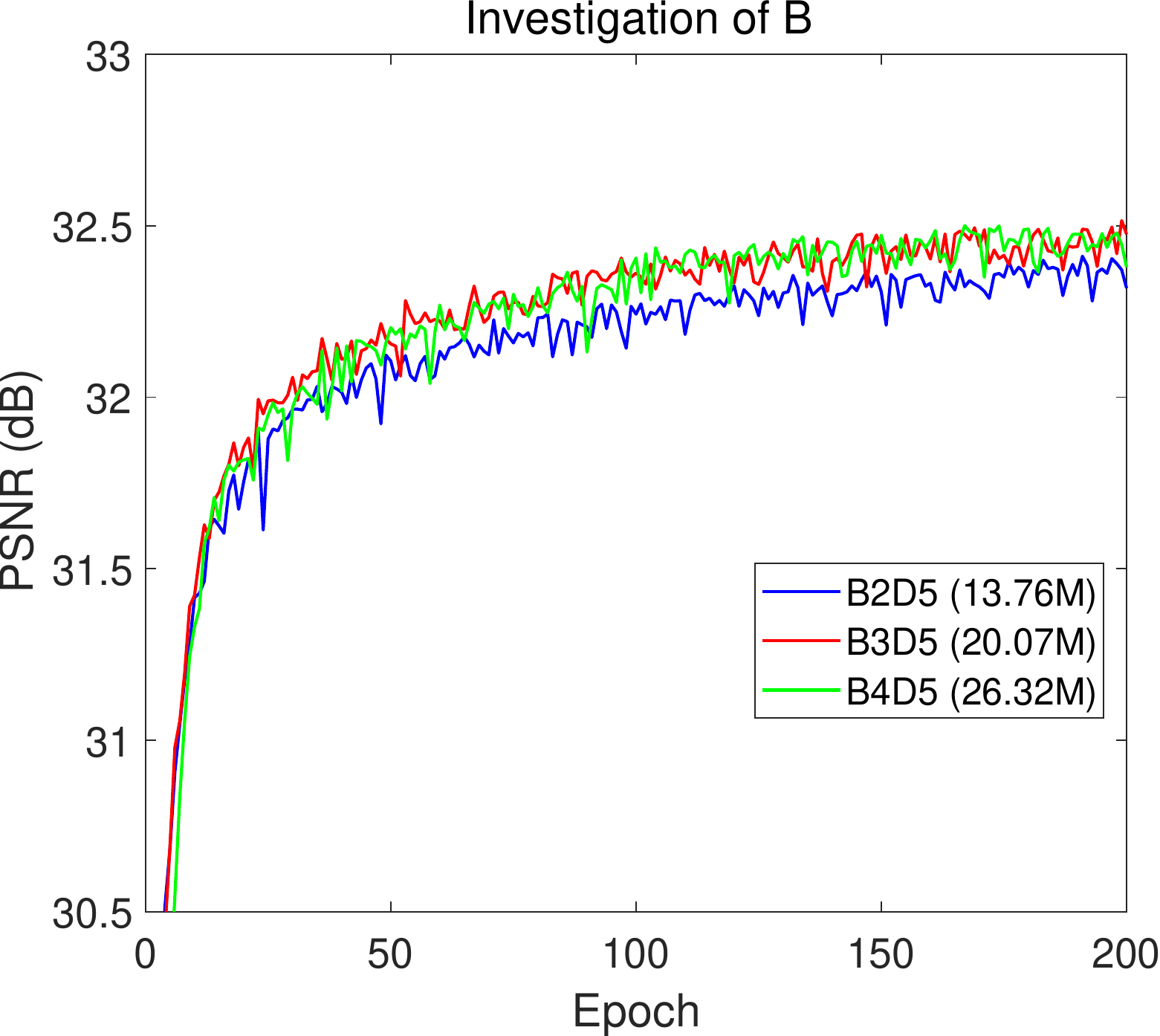}}
		\hfill
	  \caption{Convergence analysis of DIN for $4\times$ SR with different number of D and B.}
	  \label{fig6}
\end{figure}
%%%%

\subsection{Study of D and B}
In the proposed DIN, since we fix the branch number as 4, the main setting that involves network parameters is the number of WRDBs (denoted as $D$) in each branch and the number of RDBs (denoted as $B$) per WRDB. Here, we investigate the effects of two parameters for IR. As shown in Fig.~\ref{fig6:subfig:a} and \ref{fig6:subfig:b}, it can be observed that higher performance can be ensured by increase the number of $D$ or $B$. The main reason of this condition is that larger D and B can lead to deeper networks and more parameters, which demonstrates that deeper models contribute to better reconstruction results. Besides, according to the comparison in Fig.~\ref{fig6:subfig:b}, with the same number of $D=5$, when we constantly increase the number of B, the improvements from model depth encounter a bottleneck, where B4D5 shows very close performance to B3D5 but more parameters (26.32M vs. 20.07M). Therefore, for good trade-offs between performance and model complexity, we adopt $B=3$ and $D=5$  as our DIN model for further experiments.

\begin{table}[t]
    \scriptsize
    \caption{ Investigation of AsyCA, DWCs and GFF in DIN. We observe the best performance (PSNR) on Set5 with scaling factor $\times2$ in 50 epochs.}
    \label{tab2}
    \setlength{\tabcolsep}{1.8mm}
    \centering
    \begin{tabular}{|c|c|c|c|c|c|c|c|c|}
    \hline
         & \multicolumn{8}{c|}{Different combinations of AsyCA, DWCs and GFF} \\
         \hline
         \hline
         AsyCA & \XSolid & \XSolid & \Checkmark & \XSolid & \XSolid & \Checkmark & \Checkmark & \Checkmark \\ 
        \hline
        DWCs & \XSolid & \XSolid & \XSolid & \Checkmark & \Checkmark & \XSolid & \Checkmark & \Checkmark\\
        \hline  
        GFF & \XSolid & \Checkmark & \XSolid & \XSolid & \Checkmark & \Checkmark & \XSolid & \Checkmark\\
        \hline
        \hline  
        PSNR & 37.72 & 37.76 & 37.86 & 37.83 & 37.90 & 37.89 & 37.95 & 37.99\\
        \hline
    \end{tabular} 
\end{table}
%%%%

\subsection{Effectiveness of AsyCA and DWC}
We first investigate the effectiveness on the asymmetric co-attention (AsyCA) and densely weighted connections (DWCs) in WRDB. Besides, we also study the effects of the global feature fusion (GFF) strategy in DIN. As shown in Table~\ref{tab2}, the eight networks have same structure. We first train a baseline model without these three components. We then add GFF to the baseline model. In the first and second columns, when both AsyCA and DWCs are removed, the PSNR on Set5 for $2\times$ SR is relatively low, no matter the GFF is used or not. After adding one of AsyCA or DWCs to the first models, we can validate that both AsyCA and DWCs can efficiently improve the performance of networks. It can be seen that the two components respectively combined with GFF perform better than only one component adding in the GFF model. This is because that our DWCs can assign different weight parameters on different for more precise information propagation and the AsyCA can emphasize important features from different states for more discriminative feature representations. When we use these three components simultaneously, the model (the last column) achieves the best performance. These quantitative comparisons demonstrate the effectiveness and benefits of our proposed AsyCA and DWCs.

\begin{figure}[h]
\centering
\includegraphics[width=2.27in]{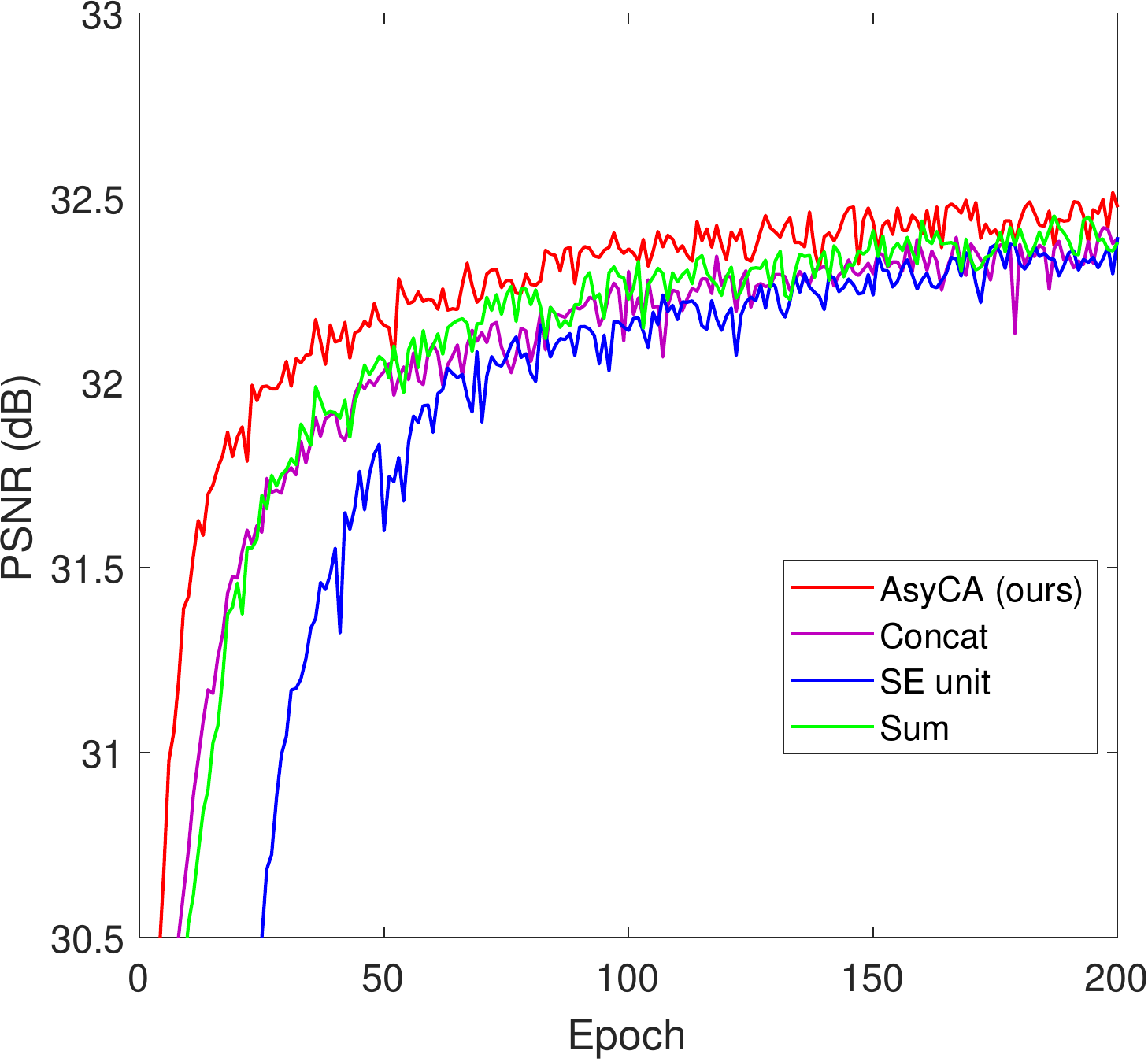}
\caption{Convergence analysis of DIN with different feature fusion
strategies: concatenation (Concat), element-wise summation (Sum), SE unit, and AsyCA, for $4\times$ SR.}
\label{fig7}
\end{figure}
%%%%

\subsection{Study of Feature Fusion Strategies}
In DIN, any feature fusion operation can be used at the interleaved nodes within the proposed IMBF. In this subsection, we compare our method with another general feature fusion methods: concatenation (denoted as Concat ), element-wise summation (denoted as sum) and SE unit~\cite{senet}. For SE based feature fusion, we first concatenate the features from two adjacent branches. Then, SE unit is used to perform channel-wise feature recalibration. We visualize the convergence process of these 4 feature fusion strategies in Fig.~\ref{fig7}. It can be seen that Concat and Sum achieve slightly higher quantitative performance than SE unit, which suggests that SE unit is not robust for feature fusion at the interleaved nodes in our framework. On the other hand, the proposed AsyCA provides better PSNR scores, lighter fluctuation, and more stable convergence process, which demonstrate the effectiveness of AsyCA for HQ images recovery.

\subsection{Model Size Analysis}
In this subsection, we compare the SR performance and model size between our DIN and recent very deep CNN-based image SR models, where the comparisons are shown in Table~\ref{tab3}. As we can see, MemNet and SRFBN contain much fewer parameters at the cost of performance degradation. Compared to EDSR and RDN, DIN achieves superior SR performance with smaller model size, especially the parameters of DIN are half less than that of EDSR. Though DIN has more parameters than SAN~\cite{san}, it suppresses SAN in terms of PSNR and SSIM by a considerable margin. The comparison in Table~\ref{tab3} demonstrate that our DIN can achieve a good trade-off between reconstruction performance and model complexity.

\begin{table}[t]
\scriptsize
%\footnotesize
%\small
%\normalsize
\centering
\begin{center}
%\vspace{-3mm}
\caption{PSNR (dB), SSIM, and parameter number (Param.) comparisons. The PSNR values are based on Set14 for $2\times$ SR.}
\label{tab3}
\vspace{-3mm}
\begin{tabular}{|@{}c@{}|c|c|c|c|c|c|c|c|c|c|c|c|c|c|c|c|c|}
%\begin{tabular}{|c|c|c|c|c|c|c|c|c|c|c|c|c|c|c|c|c|}
\hline
\multirow{2}{*}{Methods} & MemNet &  EDSR & RDN & SRFBN & SAN & DIN \\
 & \cite{memnet} & \cite{edsr} & \cite{rdn} & \cite{srfbn} & \cite{san} & (Ours)\\
%\hline
\hline
PSNR & 33.28 & 33.92 & 34.01 & 33.82 & 34.07 & 34.22                              
\\
\hline
SSIM & 0.9142 & 0.9195 & 0.9212 & 0.9196 & 0.9213 & 0.9236                              
\\
\hline  
Param. & 677K & 43M & 22M & 3.63M & 15.70M & 19.93M\\
\hline       
\end{tabular}
\end{center}
\vspace{-4mm}
\end{table}

\section{Experimental Results}
\label{sec:experiment}
In this section, we evaluate the proposed DIN and compare our models with existing state-of-the-art algorithms on several IR tasks including image SR, motion deblurring, rain streak removal, and image dehazing.

% % PSNR BI
\begin{table*}[t]
\center
\begin{center}
\caption{Quantitative results with \textbf{BI} degradation model. The best and second best results are \textbf{highlighted} and \underline{underlined}.}
\label{tab:results_BI_5sets}
\setlength{\tabcolsep}{3.14mm}{\begin{tabular}{|l|c|c|c|c|c|c|c|c|c|c|c|}
\hline
\multirow{2}{*}{Method} & \multirow{2}{*}{Scale} &  \multicolumn{2}{c|}{Set5} &  \multicolumn{2}{c|}{Set14} &  \multicolumn{2}{c|}{BSDS100} &  \multicolumn{2}{c|}{Urban100} &  \multicolumn{2}{c|}{Manga109}  
\\
%\hline
\cline{3-12}
&  & PSNR & SSIM & PSNR & SSIM & PSNR & SSIM & PSNR & SSIM & PSNR & SSIM 
\\
\hline
\hline
Bicubic & $\times$2 & 33.66 & 0.9299 & 30.24 & 0.8688 & 29.56 & 0.8431 & 26.88 & 0.8403 & 30.80 & 0.9339 \\
SRCNN~\cite{srcnn} & $\times$2 & 36.66 & 0.9542 & 32.45 & 0.9067 & 31.36 & 0.8879 & 29.50 & 0.8946 & 35.60 & 0.9663 \\
FSRCNN~\cite{fsrcnn} & $\times$2 & 37.05 & 0.9560 & 32.66 & 0.9090 & 31.53 & 0.8920 & 29.88 & 0.9020 & 36.67 & 0.9710 \\
VDSR~\cite{vdsr} & $\times$2  & 37.53 & 0.9590 & 33.05 & 0.9130 & 31.90 & 0.8960 & 30.77 & 0.9140 & 37.22 & 0.9750 \\
LapSRN~\cite{lapsrn} & $\times$2 & 37.52 & 0.9591 & 33.08 & 0.9130 & 31.08 & 0.8950 & 30.41 & 0.9101 & 37.27 & 0.9740 \\
MemNet~\cite{memnet} & $\times$2 & 37.78 & 0.9597 & 33.28 & 0.9142 & 32.08 & 0.8978 & 31.31 & 0.9195 & 37.72 & 0.9740 \\
EDSR~\cite{edsr} & $\times$2 & 38.11 & 0.9602 & 33.92 & 0.9195 & 32.32 & 0.9013 & 32.93 & 0.9351 & 39.10 & 0.9773 \\
DBPN~\cite{dbpn} & $\times$2 & 38.09 & 0.9600 & 33.85 & 0.9190 & 32.27 & 0.9000 & 32.55 & 0.9324 & 38.89 & 0.9775        
\\
RDN~\cite{rdn} & $\times$2 & 38.24 & 0.9614 & 34.01 & 0.9212 & 32.34 & 0.9017 & 32.89 & 0.9353 & 39.18 & 0.9780 \\
SRFBN~\cite{srfbn} & $\times$2 & 38.11 & 0.9609 & 33.82 & 0.9196 & 32.29 & 0.9010 & 32.62 & 0.9328 & 39.08 & 0.9779\\
SAN~\cite{san} & $\times$2 & 38.31 & 0.9620 & 34.07 & 0.9213 & 32.42 & 0.9028 & 33.10 & 0.9370 & 39.32 & 0.9792\\
DIN (ours) & $\times$2  & \underline{38.33} & \underline{0.9621} & \underline{34.22} & \underline{0.9236} & \underline{32.43} & \underline{0.9030} & \underline{33.36} & \underline{0.9388} & \underline{39.59} & \underline{0.9790} \\
DIN+ (ours) & $\times$2 & \textbf{38.38} & \textbf{0.9625} & \textbf{34.37} & \textbf{0.9245} & \textbf{32.48} & \textbf{0.9036} & \textbf{33.55} & \textbf{0.9403} & \textbf{39.74} & \textbf{0.9794} \\
\hline
\hline
Bicubic & $\times$3 & 30.39 & 0.8682 & 27.55 & 0.7742 & 27.21 & 0.7385 & 24.46 & 0.7349 & 26.95 & 0.8556 \\
SRCNN~\cite{srcnn} & $\times$3 & 32.75 & 0.9090 & 29.30 & 0.8215 & 28.41 & 0.7863 & 26.24 & 0.7989 & 30.48 & 0.9117 \\
FSRCNN~\cite{fsrcnn} & $\times$3 & 33.18 & 0.9140 & 29.37 & 0.8240 & 28.53 & 0.7910 & 26.43 & 0.8080 & 31.10 & 0.9210 \\
VDSR~\cite{vdsr} & $\times$3 & 33.67 & 0.9210 & 29.78 & 0.8320 & 28.83 & 0.7990 & 27.14 & 0.8290 & 32.01 & 0.9340 \\
LapSRN~\cite{lapsrn} & $\times$3 & 33.82 & 0.9227 & 29.87 & 0.8320 & 28.82 & 0.7980 & 27.07 & 0.8280 & 32.21 & 0.9350 \\
MemNet~\cite{memnet} & $\times$3 & 34.09 & 0.9248 & 30.00 & 0.8350 & 28.96 & 0.8001 & 27.56 & 0.8376 & 32.51 & 0.9369 \\
EDSR~\cite{edsr} & $\times$3 & 34.65 & 0.9280 & 30.52 & 0.8462 & 29.25 & 0.8093 & 28.80 & 0.8653 & 34.17 & 0.9476 \\
RDN~\cite{rdn} & $\times$3 & 34.71 & 0.9296 & 30.57 & 0.8468 & 29.26 & 0.8093 & 28.80 & 0.8653 & 34.13 & 0.9484 \\
SRFBN~\cite{srfbn} & $\times$3 & 34.70 & 0.9292 & 30.51 & 0.8461 & 29.24 & 0.8084 & 28.73 & 0.8641 & 34.18 & 0.9481\\
SAN~\cite{san} & $\times$3 & 34.75 & 0.9300 & 30.59 & 0.8476 & 29.33 & 0.8112 & 28.93 & 0.8671 & 34.30 & 0.9494 \\
DIN (ours) & $\times$3 & \underline{34.84} & \underline{0.9305} & \underline{30.71} & \underline{0.8490} & \underline{29.36} & \underline{0.8115} & \underline{29.20} & \underline{0.8726} & \underline{34.73} & \underline{0.9509} \\
DIN+ (ours) & $\times$3 & \textbf{34.90} & \textbf{0.9309} & \textbf{30.78} & \textbf{0.8502} & \textbf{29.40} & \textbf{0.8123} & \textbf{29.36} & \textbf{0.8747} & \textbf{34.93} & \textbf{0.9518} \\
\hline
\hline
Bicubic & $\times$4 & 28.42 & 0.8104 & 26.00 & 0.7027 & 25.96 & 0.6675 & 23.14 & 0.6577 & 24.89 & 0.7866 \\
SRCNN~\cite{srcnn} & $\times$4 & 30.48 & 0.8628 & 27.50 & 0.7513 & 26.90 & 0.7101 & 24.52 & 0.7221 & 27.58 & 0.8555 \\
FSRCNN~\cite{fsrcnn} & $\times$4 & 30.72 & 0.8660 & 27.61 & 0.7550 & 26.98 & 0.7150 & 24.62 & 0.7280 & 27.90 & 0.8610 \\
VDSR~\cite{vdsr} & $\times$4 & 31.35 & 0.8830 & 28.02 & 0.7680 & 27.29 & 0.0726 & 25.18 & 0.7540 & 28.83 & 0.8870 \\
LapSRN~\cite{lapsrn} & $\times$4 & 31.54 & 0.8850 & 28.19 & 0.7720 & 27.32 & 0.7270 & 25.21 & 0.7560 & 29.09 & 0.8900 \\
MemNet~\cite{memnet} & $\times$4 & 31.74 & 0.8893 & 28.26 & 0.7723 & 27.40 & 0.7281 & 25.50 & 0.7630 & 29.42 & 0.8942 \\
SRDenseNet~\cite{srdensenet} & $\times$4 & 32.02 & 0.8930 & 28.50 & 0.7780 & 27.53 & 0.7337 & 26.05 & 0.7819 & N/A & N/A \\
EDSR~\cite{edsr} & $\times$4 & 32.46 & 0.8968 & 28.80 & 0.7876 & 27.71 & 0.7420 & 26.64 & 0.8033 & 31.02 & 0.9148 \\
DBPN~\cite{dbpn} & $\times$4 & 32.47 & 0.8980 & 28.82 & 0.7860 & 27.72 & 0.7400 & 26.38 & 0.7946 & 30.91 & 0.9137 \\
RDN~\cite{rdn} & $\times$4 & 32.47 & 0.8990 & 28.81 & 0.7871 & 27.72 & 0.7419 & 26.61 & 0.8028 & 31.00 & 0.9151 \\
SRFBN~\cite{srfbn} & $\times$4 & 32.47 & 0.8983 & 28.81 & 0.7868 & 27.72 & 0.7409 & 26.60 & 0.8015 & 31.15 & 0.9160\\
SAN~\cite{san} & $\times$4 & 32.64 & 0.9003 & 28.92 & 0.7888 & 27.78 & 0.7439 & 26.79 & 0.8068 & 31.18 & 0.9169 \\
DIN (ours) & $\times$4 & \underline{32.72} & \underline{0.9011} & \underline{28.93} & \underline{0.7899} & \underline{27.81} & \underline{0.7442} & \underline{27.01} & \underline{0.8129} & \underline{31.55} & \underline{0.9195} \\
DIN+ (ours) & $\times$4 & \textbf{32.76} & \textbf{0.9016} & \textbf{29.03} & \textbf{0.7914} & \textbf{27.87} & \textbf{0.7453} & \textbf{27.17} & \textbf{0.8160} & \textbf{31.80} & \textbf{0.9214}\\
\hline
\hline
Bicubic & $\times$8 & 24.40 & 0.6580 & 23.10 & 0.5660 & 23.67 & 0.5480 & 20.74 & 0.5160 & 21.47 & 0.6500 \\
SRCNN~\cite{srcnn} & $\times$8 & 25.33 & 0.6900 & 23.76 & 0.5910 & 24.13 & 0.5660 & 21.29 & 0.5440 & 22.46 & 0.6950 \\
FSRCNN~\cite{fsrcnn} & $\times$8 & 20.13 & 0.5520 & 19.75 & 0.4820 & 24.21 & 0.5680 & 21.32 & 0.5380 & 22.39 & 0.6730 \\
VDSR~\cite{vdsr} & $\times$8 & 25.93 & 0.7240 & 24.26 & 0.6140 & 24.49 & 0.5830 & 21.70 & 0.5710 & 23.16 & 0.7250 \\   
LapSRN~\cite{lapsrn} & $\times$8 & 26.15 & 0.7380 & 24.35 & 0.6200 & 24.54 & 0.5860 & 21.81 & 0.5810 & 23.39 & 0.7350 \\
MemNet~\cite{memnet} & $\times$8 & 26.16 & 0.7414 & 24.38 & 0.6199 & 24.58 & 0.5842 & 21.89 & 0.5825 & 23.56 & 0.7387 \\
EDSR~\cite{edsr} & $\times$8 & 27.12 & 0.7798 & 24.95 & 0.6421 & 24.84 & 0.5985 & 22.55 & 0.6215 & 24.68 & 0.7826 \\
MSLapSRN~\cite{mslapsrn} & $\times$8 & 26.34 & 0.7558 & 24.57 & 0.6273 & 24.65 & 0.5895 & 22.06 & 0.5963 & 23.90 & 0.7564\\
DBPN~\cite{dbpn} & $\times$8 & 27.21 & 0.7840 & 25.13 & 0.6480 & 24.88 & 0.6010 & 22.73 & 0.6312 & 25.14 & 0.7987 \\
RDN~\cite{rdn} & $\times$8 & 27.14 & 0.7799 & 25.06 & 0.6442 & 24.86 & 0.5992 & 22.62 & 0.6239 & 24.76 & 0.7830\\
MSRN~\cite{msrn} & $\times$8 & 26.59 & 0.7254 & 24.88 & 0.5961 & 24.70 & 0.5410 & 22.37 & 0.5977 & 24.28 & 0.7517\\
SRFBN~\cite{srfbn} & $\times$8 & 27.19 & 0.7837 & 25.05 & 0.6455 & 24.86 & 0.5997 & 22.58 & 0.6256 & 24.86 & 0.7904\\
SAN~\cite{san} & $\times$8 & 27.22 & 0.7829 & 25.14 & 0.6476 & 24.88 & 0.6011 & 22.70 & 0.6314 & 24.85 & 0.7906\\
DIN (ours) & $\times$8 & \underline{27.40} & \underline{0.7903} & \underline{25.33} & \underline{0.6535} & \underline{24.98} & \underline{0.6053} & \underline{23.05} & \underline{0.6462} & \underline{25.37} & \underline{0.8045} \\
DIN+ (ours) & $\times$8 & \textbf{27.48} & \textbf{0.7934} & \textbf{25.41} & \textbf{0.6555} & \textbf{25.04} & \textbf{0.6070} & \textbf{23.20} & \textbf{0.6516} & \textbf{25.60} & \textbf{0.8092} \\
\hline             
\end{tabular}}
\end{center}
\end{table*}

\subsection{Datasets}
Here, we provide the details of experimental setup for each specific task, \emph{i.e.}, datasets and training settings. 

\textbf{Image SR}. For the training set, we use 800 HR images from DIV2K~\cite{div2k} and 2650 HR images from Flickr2K~\cite{edsr}, totally 3450 images (termed DF2K), for training. During training, we augment the training images by randomly flipping horizontally and rotating $90^{\circ}$. For testing, we evaluate our SR results on five public standard benchmark datasets: Set5~\cite{set5}, Set14~\cite{set14}, BSDS100~\cite{bsds100}, Urban100~\cite{selfexsr}, and Manga109~\cite{manga109}. All the SR results are evaluated with PSNR and SSIM on Y channel of the transformed YCbCr color space. 

\textbf{Motion Deblurring}. Following the configuration of ~\cite{msdeblur,zhang}, we train our models on GoPro dataset~\cite{msdeblur}, which consists of 3214 image pairs of blurry and sharp images at $720\times1280$ resolution. We use 2103 pairs for training and 1111 pairs for testing. Besides, we also conduct comparisons on HIDE dataset~\cite{hide} that are comprised of 1063 long-shot images (HIDE~\uppercase\expandafter{\romannumeral1}) which involves only weak independent human movement and 962 images (HIDE~\uppercase\expandafter{\romannumeral2}) that focuses on foreground moving human. PSNR and SSIM are adopted for quantitative comparison.

\textbf{Rain Streak Removal}. For image deraining, we conduct comparisons on 3 synthetic benchmarks: Rain100L~\cite{djr} , Rain100H~\cite{djr}, Rain12\footnote{\url{http://yu-li.github.io/}}~\cite{rain12}, and Rain1400~\cite{ddn}. Rain100H contains 1800 pair images for training and 100 for testing. Rain100L consists of 200 training images and 100 testing images. Noticed that the models for Rain100L and Rain100H inference are trained respectively and the Rain100L model is directly employed for Rain12 testing. For Rain1400, according to \cite{ddn}, the model are trained on 12600 rainy images and evaluated on another 1400 images. PSNR and SSIM are used as comparison criteria.
%%%%
\begin{table*}[t]
\centering
\begin{center}
\caption{Benchmark results with \textbf{BD} and \textbf{DN} degradation models. Average PSNR/SSIM values for scaling factor $\times3$. The best and second best results are \textbf{highlighted} and \underline{underlined}.} 
\label{tab:results_BD_DN_5sets}
\setlength{\tabcolsep}{1.84mm}{\begin{tabular}{|c|@{}c@{}|@{}c@{}|c|c|c|c|c|c|c|}
\hline
Dataset & Model & bicubic & VDSR~\cite{vdsr}& IRCNN~\cite{ircnn} & EDSR~\cite{edsr} & RDN~\cite{rdn} & SRFBN~\cite{srfbn} & DIN (ours) & DIN+ (ours)\\
\hline
\hline
\multirow{2}{*}{Set5} & \textbf{BD} & 32.05/0.8944 & 33.25/0.9150 & 33.38/0.9182 & 34.66/0.9285 & 34.58/0.9280 & 34.66/0.9283 & \underline{34.82}/\underline{0.9298} & \textbf{34.88}/\textbf{0.9302}\\
& \textbf{DN} & 24.01/0.5369 & 25.20/0.7183 & 25.70/0.7379 & 28.58/0.8192 & 28.47/0.8151 & 28.53/0.8182 & \underline{28.68}/\underline{0.8215} & \textbf{28.72}/\textbf{0.8222}\\
\hline
\hline
\multirow{2}{*}{Set14} & \textbf{BD} & 26.38/0.7271 & 29.46/0.8244 & 29.63/0.8281 & 30.56/0.8447 & 30.53/0.8447 & 30.48/0.8439 & \underline{30.75}/\underline{0.8482} & \textbf{30.84}/\textbf{0.8494}\\
& \textbf{DN} & 22.87/0.4724 & 24.00/0.6112 & 24.45/0.6305 & 26.67/0.7144 & 26.60/0.7101 & 26.60/0.7144 & \underline{26.74}/\underline{0.7166} & \textbf{26.77}/\textbf{0.7174}\\
\hline 
\hline
\multirow{2}{*}{BSDS100} & \textbf{BD} & 26.33/0.6918 & 28.57/0.7893 & 28.65/0.7922 & 29.25/0.8078 & 29.23/0.8079 & 29.21/0.8069 & \underline{29.39}/\underline{0.8113} & \textbf{29.43}/\textbf{0.8119}\\
& \textbf{DN} & 22.92/0.4449 & 24.00/0.5749 & 24.28/0.5900 & 25.98/0.6619 & 25.93/0.6573 & 25.95/0.6625 & \underline{26.02}/\underline{0.6633} & \textbf{26.04}/\textbf{0.6640}\\ 
\hline
\hline
\multirow{2}{*}{Urban100} & \textbf{BD} & 23.52/0.6862 & 26.61/0.8136 & 26.77/0.8154 & 28.54/0.8593 & 28.46/0.8582 & 28.48/0.8581 & \underline{29.11}/\underline{0.8700} & \textbf{29.26}/\textbf{0.8722} \\
& \textbf{DN} & 21.63/0.4687 & 22.22/0.6096 & 22.90/0.6429 & 25.09/0.7441 & 24.92/0.7364 & 24.99/0.7424 & \underline{25.29}/\underline{0.7512} & \textbf{25.36}/\textbf{0.7534}\\
\hline
\hline
\multirow{2}{*}{Manga109} & \textbf{BD} & 25.46/0.8149 & 31.06/0.9234 & 31.15/0.9245 & 34.14/0.9470 & 33.97/0.9465 & 34.07/0.9466 & \underline{34.87}/ \underline{0.9506} & \textbf{35.07}/\textbf{0.9515}\\
& \textbf{DN} & 23.01/0.5381 & 24.20/0.7525 & 24.88/0.7765 & 28.14/0.8624 & 28.00/0.8591 & 28.02/0.8618 & \underline{28.37}/\underline{0.8670} & \textbf{28.47}/\textbf{0.8687}\\
\hline

\end{tabular}}
\end{center}
\end{table*}
%%%%

%%%%
\begin{figure*}[t]
\centering
\includegraphics[width=7.11in]{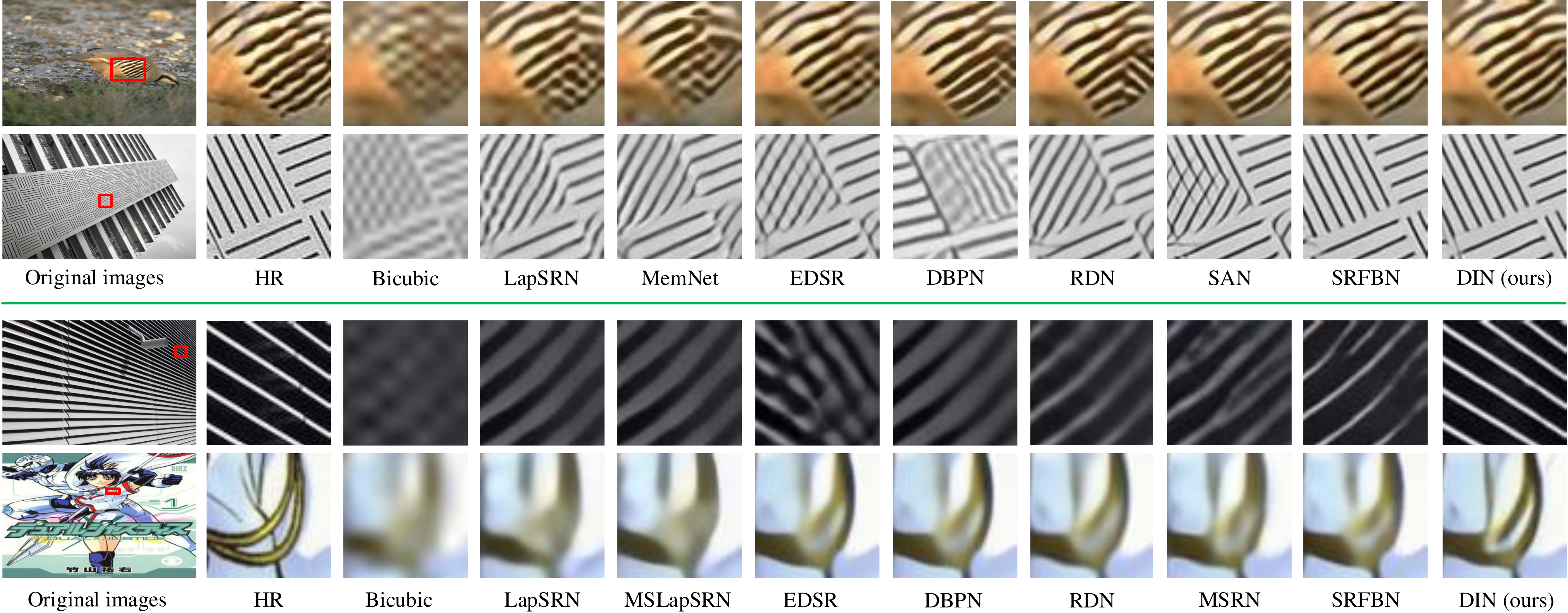}
\caption{Visual comparisons (\textbf{BI} degradation) with scaling factors $\times4$ (first two rows) and $\times8$ (last two rows).}
\label{fig8}
\end{figure*}
%%%%

%%%%
\begin{figure*}[t]
\centering
	  \subfigure[Visual comparisons (\textbf{BD} degradation)]{\label{fig9:subfig:a} 
       \includegraphics[width=3.52in]{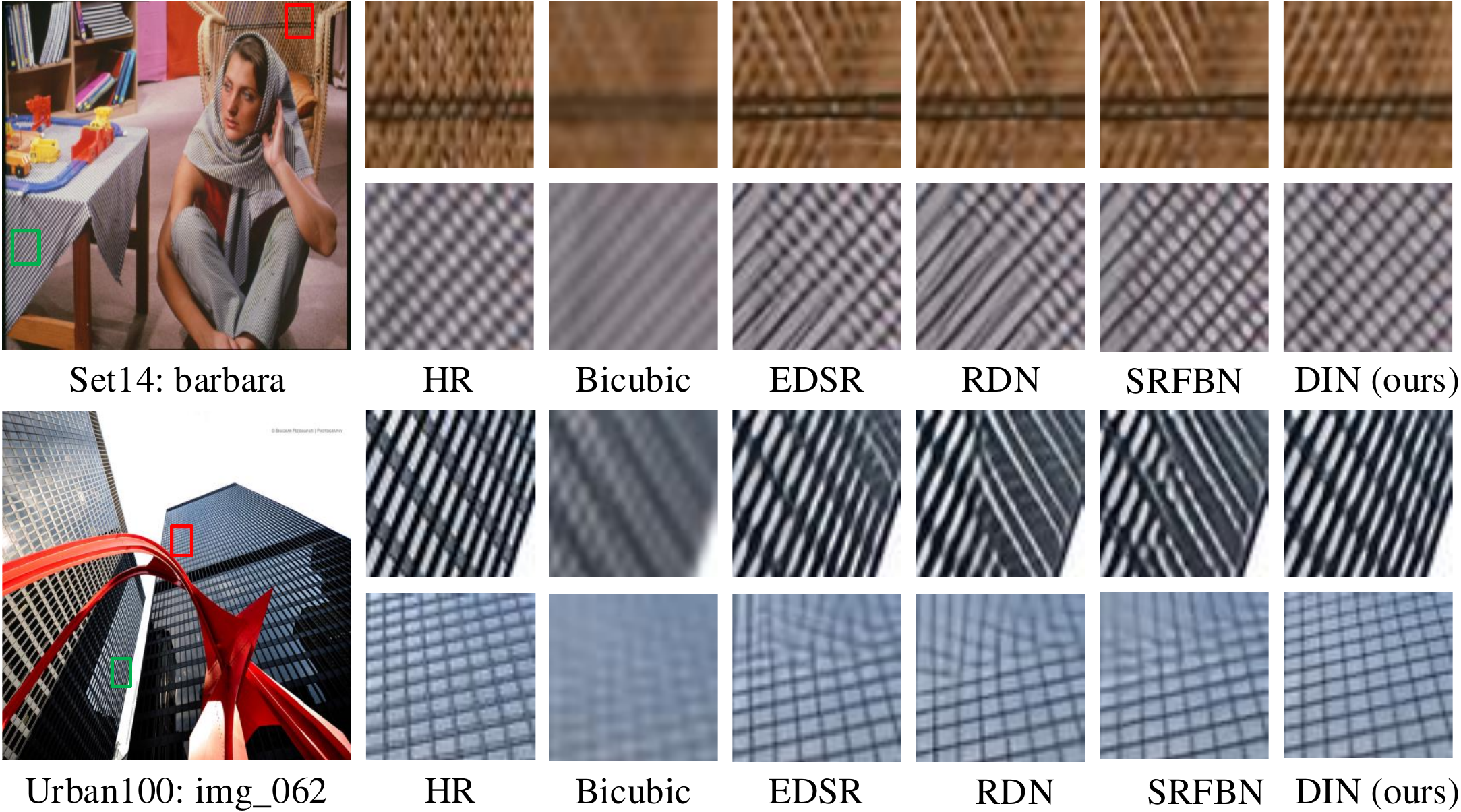}}
		\hfill
	  \subfigure[Visual comparisons (\textbf{DN} degradation)]{\label{fig9:subfig:b}
        \includegraphics[width=3.52in]{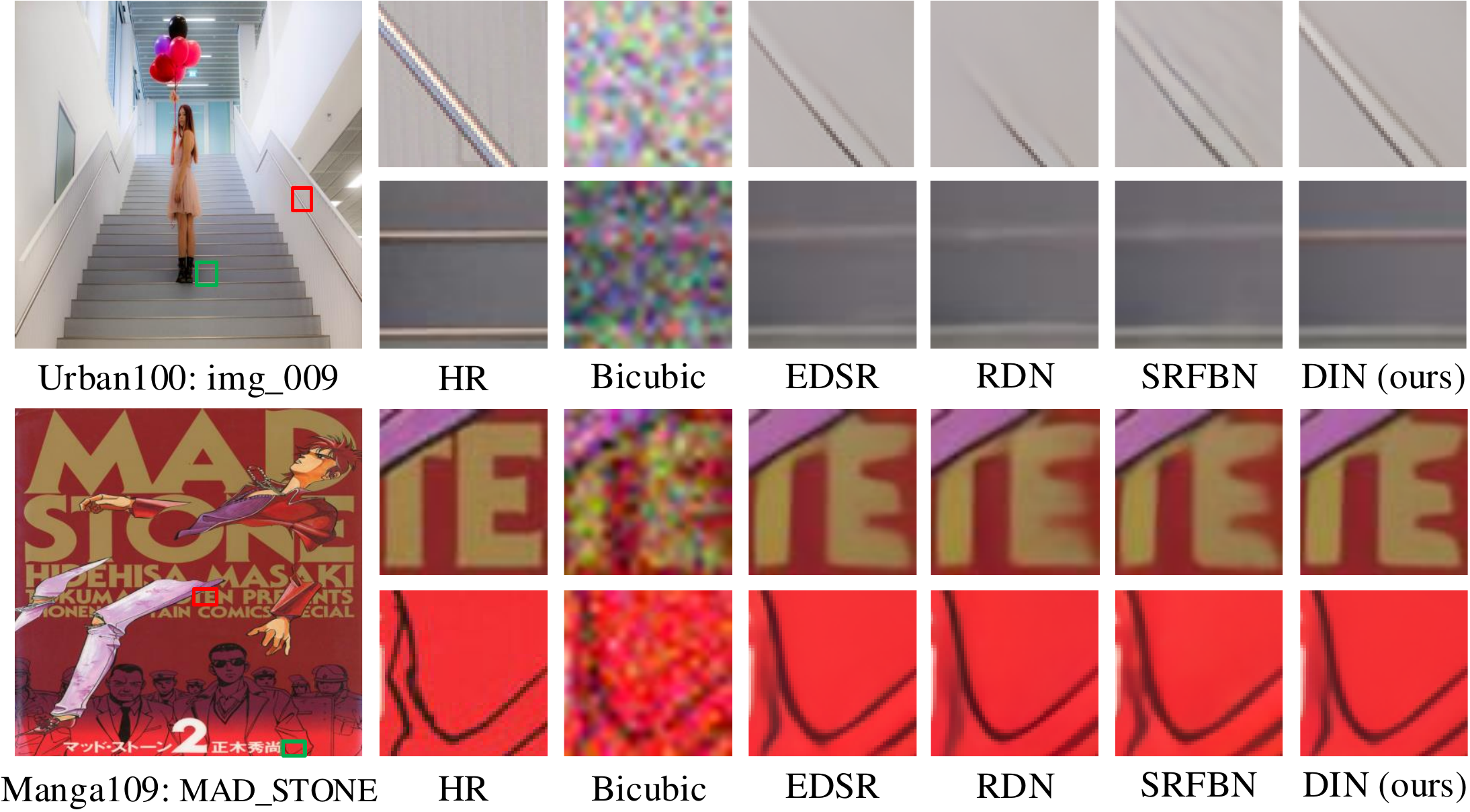}}
		\hfill
	  \caption{Visual comparisons of (a) \textbf{BD} degradation and (b) \textbf{DN} degradation with scaling factor $\times3$.}
	  \label{fig9}
\end{figure*}
%%%%

\textbf{Image Dehazing}. In this paper, we use the large-scale synthetic dehazing dataset in RESIDE~\cite{reside}, \emph{i.e.}, Indoor Training Set (ITS) and Outdoor Training Set (OTS), to train our DIN for haze removal. The ITS contains 13990 hazy indoor images, generated from 1399 clear images and OTS consists of 296695 outdoor hazy images, generated from 8477 clear ones. We adopt the Synthetic Objective Testing Set (SOTS) that is composed of  500 indoor and outdoor hazy images for testing. Besides, real-world evaluation is also conducted on the dataset from \cite{realworldhaze}.

\subsection{Training}
For image SR and deraining, in each min-batch, we randomly extract $48\times48$ patch from LR images as input for training. With respect to dehazing and motion blur removal, we randomly crop every input image to a $256\times256$ patch for each iteration. All the models are trained by Adam optimizer~\cite{adam} with $\beta_1=0.9$, $\beta_2=0.99$, and $\epsilon=10^{-8}$. We used a batch size of 8 and implement our networks with Pytorch framework on Nvidia Titan RTX GPUs.

\subsection{Comparing with the State-of-the-arts}
\subsubsection{Image Super-resolution}
In order to demonstrate the effectiveness of our proposed method, we conduct the experiments on three degradation models. The first one is the bicubic degradation model (\textbf{BI}), which uses the bicubic downsampling to generate LR images from ground truth HR images with scaling factor $\times2$, $\times3$, $\times$4 and $\times8$. The second one, blur-downscale degradation (termed as \textbf{BD}), is to blur HR image by $7\times7$ Gaussian kernel of standard deviation 1.6 and then downsample the burred image with scaling factor $\times3$. The third one is bicubic downsampling followed by adding Gaussian noise (\textbf{DN}), which conducts bicubic downsampling on HR image with scaling factor $\times3$ and then add Gaussian noise with noise level 30. Self-ensemble strategy~\cite{edsr} is further employed to improve our DIN (denoted as DIN+).

%%%%
\begin{figure}[t]
\centering
\includegraphics[width=3.36in]{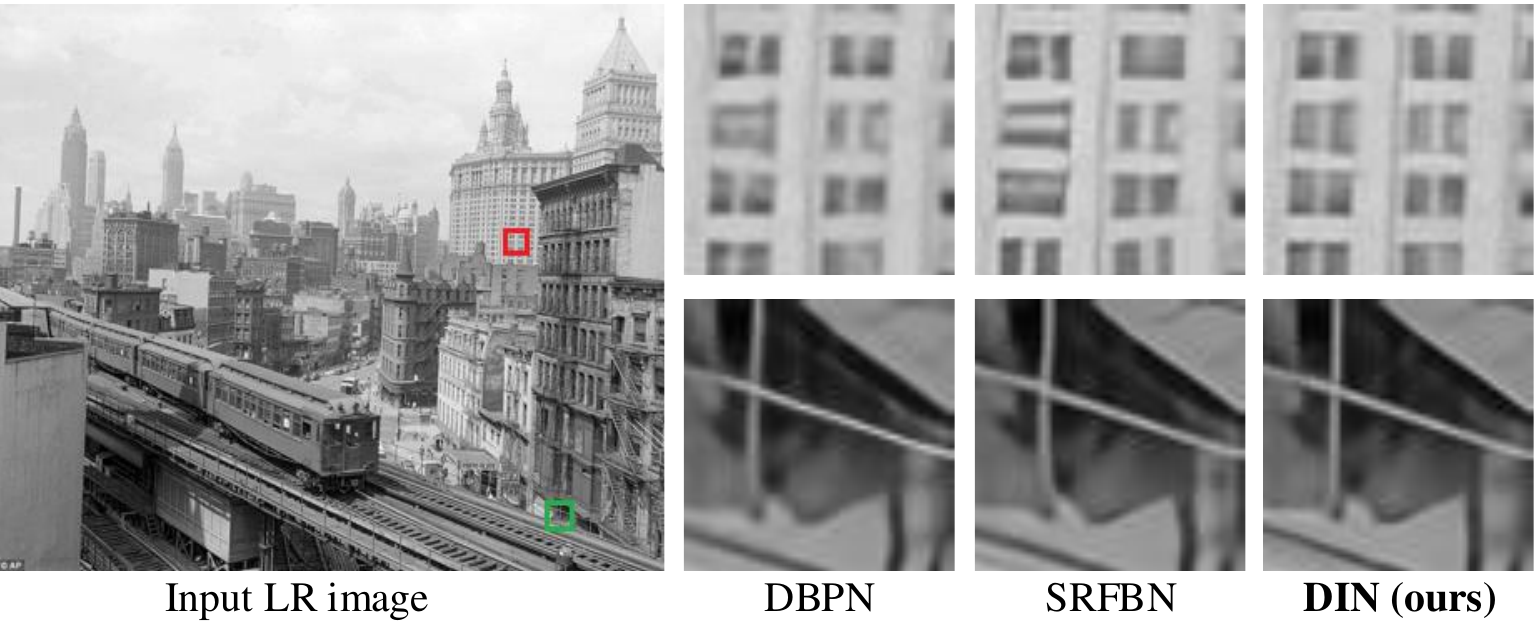}
\caption{Comparison of the real-world historical photograph for $4\times$ SR.}
\label{fig11}
\end{figure}
%%%%

\textbf{BI Degradation Model}. We compare our method with 13 CNN-based methods consisting of SRCNN~\cite{srcnn}, FSRCNN~\cite{fsrcnn}, VDSR~\cite{vdsr}, LapSRN~\cite{lapsrn}, MemNet~\cite{memnet}, EDSR~\cite{edsr}, SRDenseNet~\cite{srdensenet}, MSLapSRN~\cite{mslapsrn}, MSRN~\cite{msrn}, DBPN~\cite{dbpn}, RDN~\cite{rdn}, SRFBN~\cite{srfbn}, and SAN~\cite{san}. Since EDSR, RDN and SRFBN did not report $8\times$ SR results, we retrain their models according to the settings mentioned in their published manuscripts. The quantitative comparisons on 5 benchmark datasets are shown in Table~\ref{tab:results_BI_5sets}. Obviously, compared with other methods, DIN+ performs the best results on all the datasets for various scaling factors SR. Besides, even without the self-ensemble strategy, the proposed DIN outperforms all methods in terms of both PSNR and SSIM, especially RDN and EDSR that involve much more parameters than ours. 

In Fig.~\ref{fig8}, we show the visual comparison on $4\times$ and $8\times$ SR. In general, compared to existing methods, the proposed DIN achieves the best high-frequency details recovery capacity and yields more convincing structures and textures. Specifically, For the $4\times$ SR results of the ``8023'' from BSDS100, most methods tend to produce HR image with severely deformed lines. Though SAN and SRFBN can generate images with correct structure, the local textures still remain visible blurs. For the ``img\_092'' from Urban100, the results of almost all compared methods are suffered from wrong texture directions and serious artifacts. In contrast, our DIN can produce HR images with more realistic textures and fewer artifacts. When super-resolving LR images on a larger scale ($8\times$), such as the ``image\_40'' from Urban100, it can be observed whether bicubic or many state-of-the-art methods (\emph{e.g.} DBPN, RDN, and SRFBN) lose information fidelity and produce wrong structures, whereas our DIN can recover them correctly. For the image ``DualJustice" from Manga109, most methods show very limited ability to recover structural and high-frequency details. While, our DIN can reconstruct more pleasant results with informative context. 

\textbf{BD and DN Degradation Models}. Following \cite{rdn,san}, we additionally conduct our experiments on more challenging degradations, \emph{i.e.}, \textbf{BD} and \textbf{DN}. Our DIN are compared with VDSR~\cite{vdsr}, IRCNN~\cite{ircnn}, EDSR~\cite{edsr}, RDN~\cite{rdn}, and SRFBN~\cite{srfbn}. All of the results for $3\times$ SR are illustrated in Table~\ref{tab:results_BD_DN_5sets}, from which we can observe that the proposed DIN and DIN+ perform the best on all 5 datasets over other state-of-the-art algorithms. For qualitative comparisons, we show the visual results of \textbf{BD} and \textbf{DN} degradations in Fig.~\ref{fig9}. One can see that, for \textbf{BD} degradation, all compared algorithms suffer from critical blurs and wrong textures, whereas our DIN can generate accurate and high perceptual quality results without unpleasant blurry contents. On the other hand, for \textbf{DN} degradation, as shown in Fig~\ref{fig9:subfig:b}, conventional bicubic method can only increase the resolution of the image, but can not effectively remove the noise in corrupted images. Though EDSR and SRFBN can perform SR while reducing the noise to a certain extent, the reconstructed HR images still show worse perceptual quality than ours.

\textbf{Super-resolving Real-world Images}. To investigate the robustness of the proposed DIN, we conduct experiments on super-resolving historical photographs with JPEG compression artifacts. In this case, neither the degradation manner nor ground-truth are available.  We compare our method with two state-of-the--art methods DBPN~\cite{dbpn} and SRFBN~\cite{srfbn} on $4\times$ SR. As shown in Fig.~\ref{fig11}, our DIN can accurately super-resolve the outlines of windows (top red), straight and clear railings, while the results of DBPN and SRFBN contain distorted local image details and severe artifacts. These comparisons demonstrate the effectiveness of our method for image SR. 

\subsubsection{Motion Deblurring}
\textbf{Comparing with the State-of-the-arts}. We compare the proposed DIN with recently motion blur removal methods: Sun~\emph{et al.}~\cite{sun}, Nah~\emph{et al.}~\cite{msdeblur}, Deblurgan~\cite{deblurgan}, Tao~\emph{et al.}~\cite{srn}, and Li~\emph{et al.}~\cite{li}. As shown in Table~\ref{tab6}, in terms of PSNR/SSIM, our DIN performs favorably against with the state-of-the-art algorithms on GoPro dataset and achieves the best performance on HIDE dataset. Besides, to demonstrate the efficiency of the proposed method, we compare the computation cost by calculating the total Mult-Adds operations. The Multi-Adds of all methods are calculated by assuming the input blurred image is of 720p ($1280\times720$). As we can see, though DeblurGAN shows the most efficient performance in term of computation cost, it is at the expense of reconstruction quality. Our DIN achieves the second best results with 1198.05GFLOPs, which is fewer than Nah~\emph{et al.}~\cite{msdeblur} (1760.04GFLOPs) and Tao~\emph{et al.}~\cite{srn} (1434.82GFLOPs) by a considerable margin. The comparisons in Table~\ref{tab6} demonstrate that the proposed algorithm can achieve optimal balance between restoration performance and efficiency.

\begin{table}[t]
\scriptsize
\centering
\begin{center}
\caption{Performance on GoPro and HIDE test dataset, All models were tested on the linear image subset. The best and second best results are \textbf{highlighted} and \underline{underlined}.}
\label{tab6}
\vspace{-3mm}
\begin{tabular}{@{}c@{}|c|c|c|@{}c@{}}
%\begin{tabular}{|c|c|c|c|c|c|c|c|c|c|c|c|c|c|c|c|c|}
\hline
\multirow{3}{*}{Methods} & \multirow{2}{*}{GoPro~\cite{msdeblur}} & \multicolumn{2}{c|}{HIDE~\cite{hide}} & \multirow{2}{*}{Mult-Adds} \\
\cline{3-4}
& & HIDE \uppercase\expandafter{\romannumeral1} & HIDE \uppercase\expandafter{\romannumeral2} & \\
\cline{2-5}
& PSNR/SSIM & PSNR/SSIM & PSNR/SSIM & FLOPs\\
\hline
\hline
Sun~\emph{et al.}~\cite{sun} & 24.64/0.843 & 23.21/0.797 & 22.26/0.796 & N/A\\
Nah~\emph{et al.}~\cite{msdeblur} & 28.94/0.880 & 27.43/0.902 & 26.18/0.878 & 1760.04G\\
DeblurGAN~\cite{deblurgan} & 28.71/0.836 & 26.44/0.890 & 27.35/0.907 & 678.29G\\
Tao~\emph{et al.}~\cite{srn} & 30.26/0.934 & 28.60/0.928 & 25.37/0.867 & 1434.82G\\
Li~\emph{et al.}~\cite{li} & \textbf{30.49}/\textbf{0.938} & N/A & N/A & N/A\\
DIN~(ours) & \underline{30.43}/\textbf{0.938} & \textbf{29.48}/\textbf{0.919} & \textbf{28.00}/\textbf{0.902} & 1198.05G\\
\hline
\hline  
\end{tabular}
\end{center}
\end{table}

%%%%
\begin{figure}[t]
\centering
\includegraphics[width=3.35in]{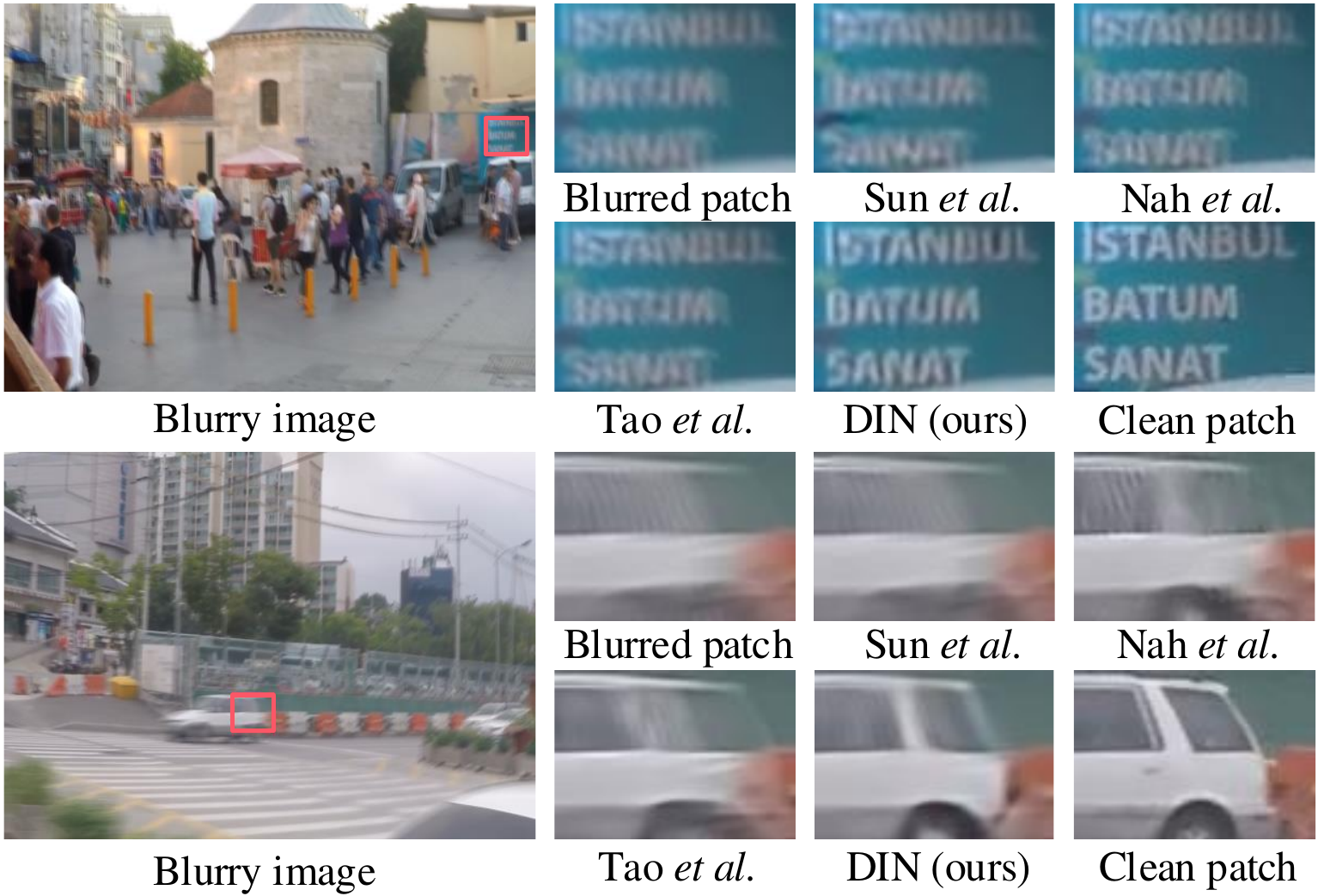}
\caption{Quantitative evaluations on GoPro dataset~\cite{msdeblur}.}
\label{fig12}
\end{figure}
%%%%

%%%%
\begin{figure}[t]
\centering
\includegraphics[width=3.50in]{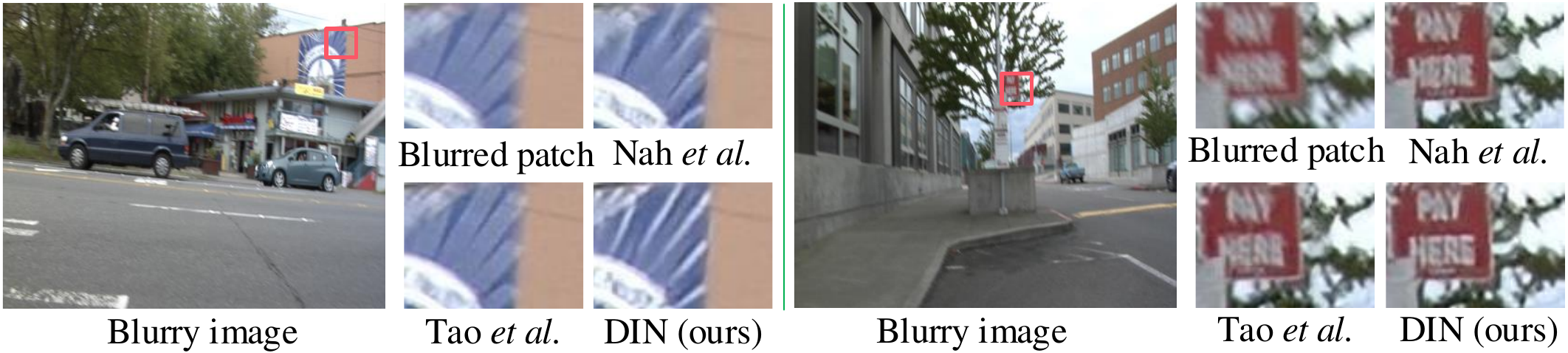}
\caption{Deblurred results on real blurry images.}
\label{fig13}
\end{figure}
%%%%

We further show the visual comparisons of deblurred images from the GoPro dataset in Fig.~\ref{fig12}. The results generated by Sun~\emph{et al.}~\cite{sun} and Nah~\emph{et al.}~\cite{msdeblur} suffer from heavy ringing artifacts and blurs. Tao~\emph{et al.}~\cite{srn} adopt scale-recurrent network to remove the motion blurs, however, there is limited performance improvement than the former two methods. In contrast, our method can better handle the strong blur on background and moving objects and produce deblurred images with clearer details.

\textbf{Real-world Motion Blurs Removal}. We further conduct experiments on the real-world video blurry dataset (720p) collected by Su~\emph{et al.}~\cite{su} to study the effectiveness of our DIN. Fig.~\ref{fig13} shows the qualitative comparisons between our DIN and existing state-of-the-art methods. It can be observed that the compared methods fail to remove blur while preserving clear details. In contrast, the proposed DIN can restore the corrupted images with more clearer textures and fewer artifacts, which demonstrates the robustness of DIN.

%%%%
\begin{table}[t]
\scriptsize
\centering
\begin{center}
\caption{PSNR and SSIM comparisons on three benchmark datasets. The best and second best results are \textbf{highlighted} and \underline{underlined}.}
\label{tab7}
\vspace{-3mm}
\begin{tabular}{|@{}c@{}|c|c|c|c|}
%\begin{tabular}{|c|c|c|c|c|c|c|c|c|c|c|c|c|c|c|c|c|}
\hline
\multirow{2}{*}{Methods} & Rain100L & Rain100H & Rain12 & Rain1400 \\
\cline{2-4}
& PSNR/SSIM & PSNR/SSIM & PSNR/SSIM & PSNR/SSIM \\
\hline
\hline
Rainy input & 26.90/0.838 & 13.56/0.379 & 30.14/0.856 & 25.24/0.810\\
\hline
GMM~\cite{gmm} & 28.66/0.865 & 15.05/0.425 & 32.02/0.855 & 27.78/0.859\\
\hline
DDN~\cite{ddn} & 32.16/0.936 & 21.92/0.764 & 31.78/0.900 & 28.45/0.889\\
\hline
JORDER~\cite{djr} & 36.61/0.974 & 26.54/0.835 & 33.92/0.953 & 32.00/0.935\\
\hline
RESCAN~\cite{rescan} & 38.52/0.981 & 29.62/0.872 & 36.43/0.952 & 32.03/0.932\\
\hline
PReNet~\cite{prenet} & \underline{37.48}/\underline{0.979} & \underline{29.46}/\underline{0.899} & \underline{36.66}/\textbf{0.961} & \underline{32.45}/\underline{0.944}\\
\hline
DIN~(ours) & \textbf{40.00}/\textbf{0.986} & \textbf{31.24}/\textbf{0.914}& \textbf{37.01}/\underline{0.954} & \textbf{33.91}/\textbf{0.955}\\
\hline  
\end{tabular}
\end{center}
\end{table}

%%%%
\begin{figure}[t]
\centering
\includegraphics[width=3.5in]{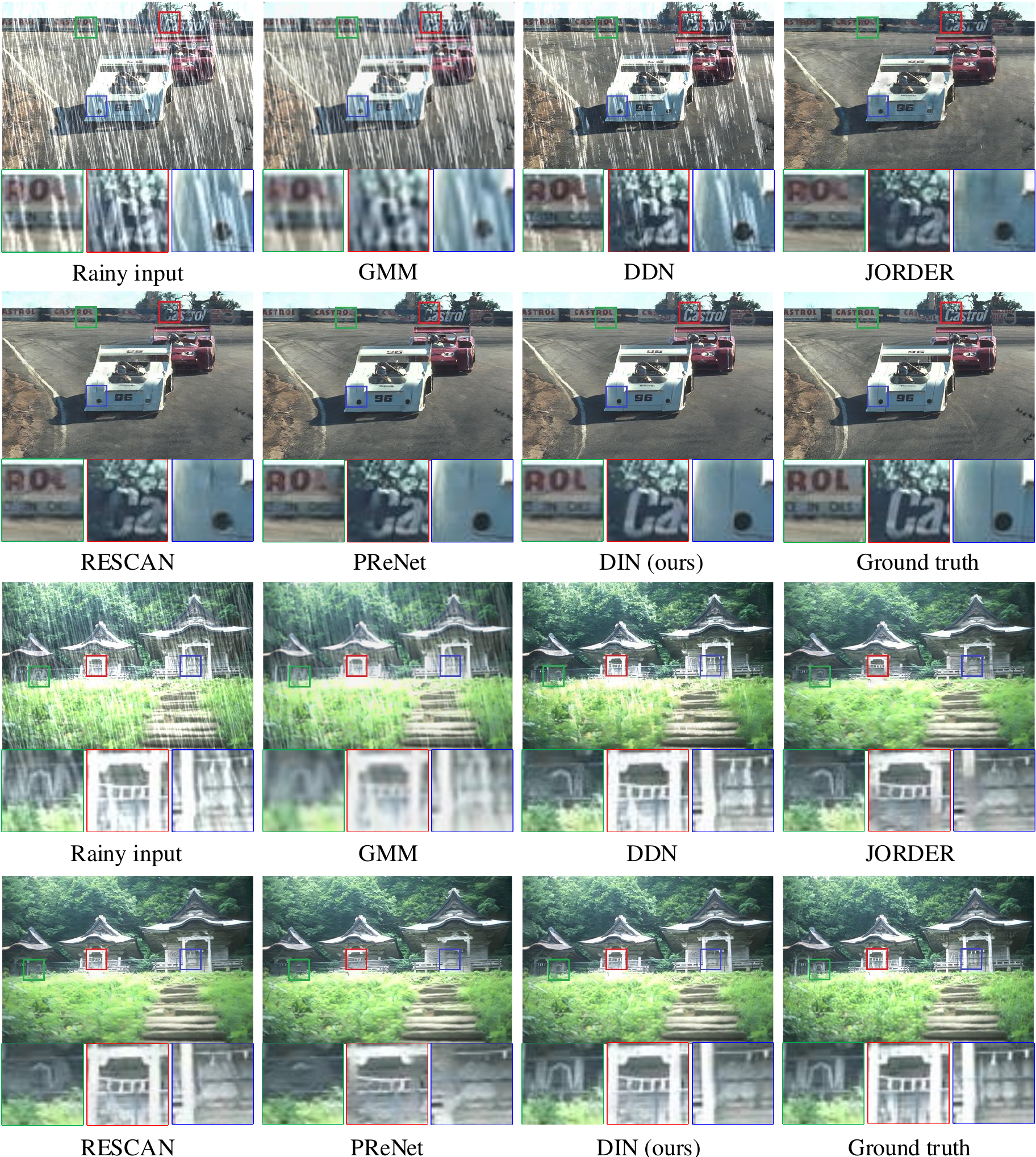}
\caption{Visual quality comparison on Rain100H.}
\label{fig16}
\end{figure}
%%%%

\subsubsection{Single Image Rain Streak Removal}
We further apply our DIN to handle image deraining problem. Here, we evaluate the proposed method against the several representative single image deraining methods including GMM\cite{gmm}, DDN~\cite{ddn}, RESCAN~\cite{rescan}, PReNet~\cite{prenet}, and SPANet~\cite{spanet}. The quantitative comparisons on 3 benchmark datasets Rain100L~\cite{djr}, Rain100H~\cite{djr}, and Rain12~\cite{rain12} are shown in Table~\ref{tab7}. It is seen that our DIN obtains the best objective quality among almost all methods on each dataset, especially exceeds the most state-of-the-art PReNet about 1.7dB in terms of PSNR on the heavy rain dataset Rain100H.

\begin{table*}[t]
\scriptsize
\centering
\begin{center}
\caption{Comparison results of the state-of-the-art dehazing methods on SOTS. The best and second best results are \textbf{highlighted} and \underline{underlined}.}
\label{tab8}
\vspace{-3mm}
\begin{tabular}{|c|c|c|c|c|c|c|c|c|c|}

\hline
\multirow{2}{*}{Scene} & DCP & DehazeNet & MSCNN & AOD-Net & DCPDN & GFN & EPDN & MSADN & DIN\\

& \cite{dark} & \cite{dehazenet} & \cite{mscnn} & \cite{aodnet} & \cite{dcpdn} & \cite{gfn} & \cite{epdn} & \cite{msadn} & (ours)\\
\hline
\hline
Indoor & 16.62/0.8179 & 21.14/0.8472 & 19.84/0.8327 & 19.06/0.8504 & 15.85/0.8175 & 22.30/0.8800 & \underline{25.06}/\underline{0.9232} & 24.02/0.9221 & \textbf{27.55}/\textbf{0.9423}\\
\hline
Outdoor & 19.13/0.8148 & 22.46/0.8514 & 22.06/0.9078 & 19.93/0.8449 &20.29/0.8765 & 21.55/0.8444 & 22.57/0.8630 & \underline{23.64}/\underline{0.9137} & \textbf{31.62}/\textbf{0.9812}\\
\hline
\end{tabular}
\end{center}
\end{table*}
%%%%

%%%%
\begin{figure*}[t]
\centering
\includegraphics[width=7.0in]{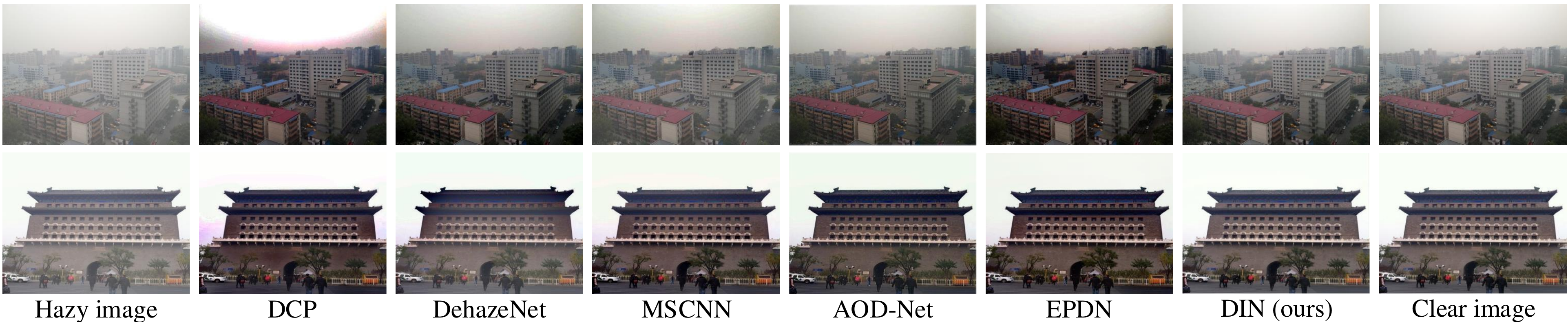}
\caption{Qualitative comparisons on SOTS outdoor scenes.}
\label{fig17}
\end{figure*}
%%%%

%%%%
\begin{figure*}[t]
\centering
\includegraphics[width=6.98in]{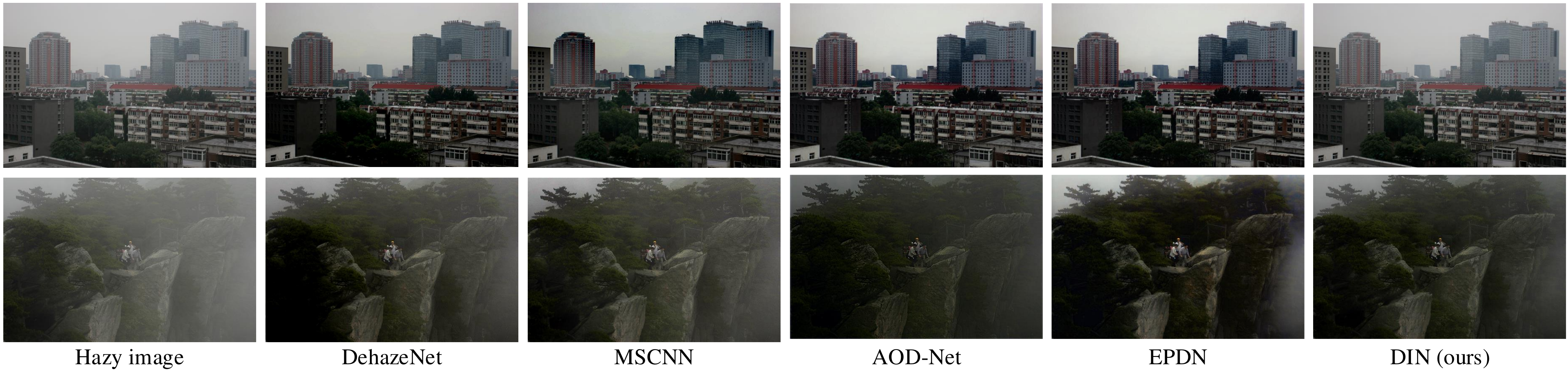}
\caption{Qualitative comparisons on real-world hazy images.}
\label{fig18}
\end{figure*}
%%%%

In Fig.~\ref{fig16}, the results of one image sample from Rain100H generated by different algorithms are given, from which we can see that some deraining methods~\cite{gmm} produce the derained image with heavy blurs. DDN~\cite{ddn} and RESCAN~\cite{rescan} can well remove the rain streaks in backgrounds but leave some visible dark noises or rain streaks in local regions. JORDER~\cite{djr} is able to remove the heavy rain streaks and produce clearer derained images than former methods. However, this method tends to miss some important structures that are similar to rain streaks. This phenomenon can also be seen in PReNet~\cite{prenet}. In comparison, DIN can effectively remove the rain streaks and generate accurate and visually promising results.

\subsubsection{Haze Removal}
\textbf{Evaluation on Synthetic Dataset}. We evaluate our methods on the Outdoor and Indoor datasets of SOTS for quantitative and qualitative evaluation with recent state-of-the-art image dehazing algorithms: DCP~\cite{dark}, DehazeNet~\cite{dehazenet}, MSCNN~\cite{mscnn}, AOD-Net~\cite{aodnet}, GFN~\cite{gfn}, DCPDN~\cite{dcpdn}, EPDN~\cite{epdn} and MSADN~\cite{msadn}. As illustrated in Table~\ref{tab8}, our method achieves the best performance on both indoor and outdoor scenes in terms of PSNR and SSIM. Especially, compared with the-state-of-the-art method EPDN, our DIN obtains the gain with about 6.5dB. Fig.~\ref{fig17} shows the visual quality comparison on synthetic dataset. As we can see, the dehazed images by DCP suffer from critical color distortions (\emph{e.g.} the sky). For the results of DehazeNet and MSCNN, there is still haze in local areas (the street of MSCNN in the 1st row and tree of DehazeNet in the 3rd row). For AOD-Net as well as EPDN, they effectively remove the haze but lead to exorbitant contrast, which seems over-dehazed. In contrast, our DIN can generate dehazed images that are more natural and visually faithful to the clear one.

\textbf{Evaluation on Real-world Images.} We further evaluate the qualitative results on real-world hazy images against the state-of-the-arts. As illustrated in Fig.~\ref{fig18}, for the image with light haze (the top two rows), all the compared methods tend to darkened the input hazy image, whereas the proposed DIN can remove the haze effectively and generate the textures with less color distortion. For the heavy hazy image (the bottom two rows), DehazeNet and MSCNN leave lots of obvious haze residuals. Thought AOD-Net and EPDN can produce better results than the former two methods, the generated images still remain severe haze artifacts and show lower brightness than ours in general. The comparisons to these state-of-the-art methods demonstrate that the proposed DIN is more effective for natural haze removal. 

\section{Conclusion}
\label{sec:conclusion}
In this paper,  a novel deep interleaved network (DIN) is proposed to handle general image restoration tasks. In the DIN, we construct a interleaved multi-branch framework (IMBF) to learn how information at different
states should be combined for high-quality image reconstruction. The main novel points of DIN lie in the built weighted residual dense block (WRDB) and asymmetric co-attention (AsyCA). The WRDB consists on multiple residual dense blocks (RDBs) and densely weighted connections (DWCs),  in which different weighted connections are assigned to different states for more precise features aggregation and propagation. The AsyCA is attached at each interleaved node among adjacent branches to adaptively emphasize the informative features from different states and generate trainable weights for feature fusion.  Extensive evaluation on public benchmarks and real-world scenes demonstrate that our DIN is more effective than state-of-the-art methods for image restoration.

\ifCLASSOPTIONcaptionsoff
  \newpage
\fi

{
\bibliographystyle{IEEEtran}
\bibliography{din}
}

\end{document}